\documentclass[letterpaper]{article} 
\usepackage[preprint]{aaai2027}  
\usepackage[hyphens]{url}  
\usepackage{graphicx} 
\usepackage{natbib}  
\usepackage{caption} 
\DeclareCaptionStyle{ruled}{labelfont=normalfont,labelsep=colon,strut=off} 
\usepackage{booktabs}
\usepackage{amsmath}
\usepackage{amssymb}
\usepackage{placeins}
\usepackage{algorithm}
\usepackage{algorithmic}

\title{Putting Registers to Work: Task Registers for Token Pruning in Vision Transformers}

\author{
Hongsen Cao,\quad Mona Jaber,\quad Shanxin Yuan,\quad Ahmed Sayed
}
\affiliations{
School of Electronic Engineering and Computer Science\\
Queen Mary University of London\\
London, United Kingdom\\
\{hong.cao, m.jaber, shanxin.yuan, ahmed.sayed\}@qmul.ac.uk
}
\begin{document}

\maketitle

\begin{abstract}
Token-pruning policies are usually designed for a single recognition pipeline, but pretrained Vision Transformers are reused across tasks with different spatial demands. We ask which parts of a pruning policy transfer across image classification, semantic segmentation, and object detection. For each pipeline, controlled probes freeze the no-pruning checkpoint and apply a series of parameter-free reduction criteria at one eligible layer at a time without retraining. The probes reveal three differences: segmentation and detection rank the criteria differently, classification is especially sensitive to attention-based pruning in the earliest layers, and the dense tasks prefer opposite recovery endpoints. These findings motivate Task-Adaptive Pruning (TAP). Existing register tokens serve as task-agnostic storage for feature artifacts. TAP instead introduces one task register per task and activates only the current one. Its evolving state ranks tokens, distributes an exact removal budget over depth, and sets the recovery scale for dense features. At a final keep rate of $\rho=0.5$, our jointly adapted model, TAP-J, reaches $47.0$ mIoU at $1.30\times$ encoder throughput on ADE20K and $53.7$ box AP at $1.32\times$ encoder throughput on COCO while remaining competitive on ImageNet-1K.
\end{abstract}

\section{Introduction}

Self-attention in a Vision Transformer (ViT) scales quadratically with the number of tokens \citep{vit,transformer}. Token pruning lowers this cost by removing low-ranked patch tokens before later blocks \citep{dynamicvit,evit,avit}. Most pruning criteria are developed for image classification, yet pretrained ViT backbones are now routinely adapted to segmentation and detection \citep{vitdet,segformer}. Applying the same pruning policy across these pipelines assumes that token importance transfers across tasks. When this assumption fails, equal token budgets need not preserve equal amounts of task-relevant information. Existing methods use fixed, learned, or task-specific selectors \citep{dynamicvit,evit,tome,tokencropr,vltp}, but generally treat pruning as a single policy. Across tasks, however, at least three choices may vary: which tokens to retain, how to distribute a global budget over depth, and how to represent removed positions for dense prediction. These decisions are often intertwined, making it difficult to see what actually transfers.

\begin{figure*}[t]
\centering
\includegraphics[width=\textwidth]{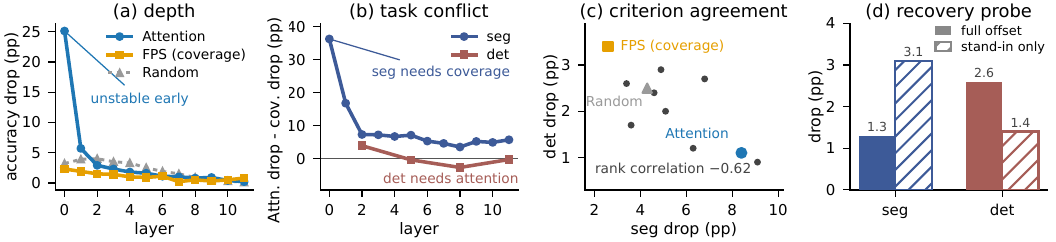}
\caption{Controlled probes apply pruning to one layer of the corresponding fixed no-pruning checkpoint and vary one design choice at a time. Attention and coverage (FPS) are highlighted as representative criteria, with random selection as a reference. Colors identify the highlighted criteria in (a) and (c), and the tasks in (b) and (d). Drops are reported in percentage points (pp). (a) For classification, attention incurs a large accuracy drop in the early layers, while coverage is more stable. (b) Attention drop minus coverage drop for segmentation and detection. Positive values favor coverage and negative values favor attention. Detection is evaluated only at global-attention blocks. (c) Each point represents an evaluated criterion and shows its mean segmentation and detection drops over the eligible layers. The rank correlation is computed across criteria. (d) Recovery with the full stored offset (solid) or the final stand-in alone (hatched).}
\label{fig:insight}
\end{figure*}

Our experiments in Figure~\ref{fig:insight} show that the effectiveness of a pruning criterion changes across network depths and tasks. We highlight two representative criteria, attention and coverage, with random selection included as a reference. For the classification task, Figure~\ref{fig:insight}(a) shows that attention-based selection causes large drops in the earliest layers of the classification pipeline, but a spatially distributed criterion (coverage) is more stable, where the coverage criterion is implemented with Farthest Point Sampling (FPS)~\citep{fps}.
For the segmentation and detection tasks, Figure~\ref{fig:insight}(b) shows that the relative advantage of attention and coverage changes with depth and differs between the two tasks. These observations motivate learning the layer allocation rather than applying the same removal schedule at every layer. Figure~\ref{fig:insight}(c) reveals a negative rank correlation between segmentation and detection tasks. The criterion with the smallest detection drop ranks among the poorer choices for segmentation, indicating a score calibrated for one dense pipeline may transfer poorly to another. As dense prediction requires a feature at every removed position, we pair each removed token with a retained stand-in and store their feature difference. We find that restoring this difference improves the segmentation task only, while using the stand-in alone improves the detection task, see (Figure~\ref{fig:insight}(d)). This motivates a recovery scale conditioned on task and depth.

Together, these findings motivate our Task-Adaptive Pruning (TAP), which puts registers to work as pruning controllers. Existing register tokens absorb feature artifacts but remain task agnostic and do not control pruning~\citep{registers}. TAP instead introduces one initial task register per task and activates only the current one during a forward pass. The active register evolves alongside the image tokens throughout the backbone. Its current state coordinates token selection, the distribution of an exact global removal budget across pruning layers, and recovery scaling for dense prediction. We study two adaptation regimes. TAP-J jointly fine-tunes a separate backbone and register for each task, while TAP-F freezes a shared backbone and trains a task-specific register together with low-rank updates. Our contributions are threefold:

\begin{enumerate}
\renewcommand{\labelenumi}{\arabic{enumi})}

\item We develop a controlled framework that freezes each no-pruning pipeline and changes one pruning choice at one eligible layer to study ViT backbone pruning across image classification, semantic segmentation, and object detection. The probes show that segmentation and detection rank pruning criteria differently, spatial coverage is more stable than attention in the early classification layers, and the dense pipelines favor opposite recovery endpoints.

\item We introduce task registers as active, task-specific controllers of sparse computation rather than task-agnostic stores for feature artifacts. Only the current task's register enters the backbone, and its evolving state coordinates token selection, exact budget allocation across depth, and dense-recovery scaling.

\item TAP introduces a task-conditioned pruning process that treats token removal and feature recovery as a unified operation, rather than relying solely on token removal or in-backbone merging. Each discarded token is matched to a surviving stand-in, and the task register scales its stored feature offset only at the dense readout, keeping reconstructed features out of later encoder blocks.

\end{enumerate}

\section{Related Work}

\subsubsection{Token reduction criteria.}
Several pruning methods rank tokens using a single criterion and discard those with low scores \citep{spvit,zerotprune}. EViT~\citep{evit} retains tokens that receive high attention from the class token, while DynamicViT~\citep{dynamicvit} predicts token retention probabilities with lightweight modules inserted between transformer blocks. Evo-ViT~\citep{evovit} maintains slow and fast token sets, and ATS~\citep{ats} samples tokens according to their attention scores. Token merging shortens the sequence by combining tokens \citep{tokenpooling}. ToMe~\citep{tome} uses bipartite matching, while DiffRate~\citep{diffrate} learns layerwise compression rates. TPS fuses information from pruned tokens into the retained sequence for classification~\citep{tps}. ALGM~\citep{algm} and SegFormer++~\citep{segformerpp} adapt token merging to semantic segmentation. In contrast to these in-backbone reduction schemes, TAP keeps the encoder sparse after removal and reconstructs dense positions only at downstream read points.

\subsubsection{Task-aware and diversity-based pruning.}
BAT~\citep{bat} preserves attentive tokens while merging similar inattentive tokens to maintain diversity, a goal related to the coverage criterion in our probe. Token Cropr~\citep{tokencropr} learns task-relevant scores through auxiliary prediction heads that are removed after training. VLTP~\citep{vltp} conditions token relevance on vision-language guidance for task-oriented segmentation. Dynamic Tuning~\citep{dyt} pairs lightweight adapters with a token dispatcher that allows less informative tokens to skip the original transformer block. TAP differs from these approaches by assigning one register to each task and using the active register to jointly condition token selection, layerwise budget allocation, and recovery scaling.

\subsubsection{Controlled studies of token reduction.}
\citet{haurum} evaluate ten methods on four image-classification datasets. For dense prediction, DToP \citep{dtop} introduces token early exit for semantic segmentation, while SViT \citep{svit} studies token preservation and reactivation for detection and instance segmentation. Sparse DETR \citep{sparsedetr} learns sparsity within a detection pipeline. Our probes hold each task pipeline fixed and ask how the same criteria rank across three settings.

\section{Method}

\begin{figure*}[t]
\centering
\includegraphics[width=\textwidth]{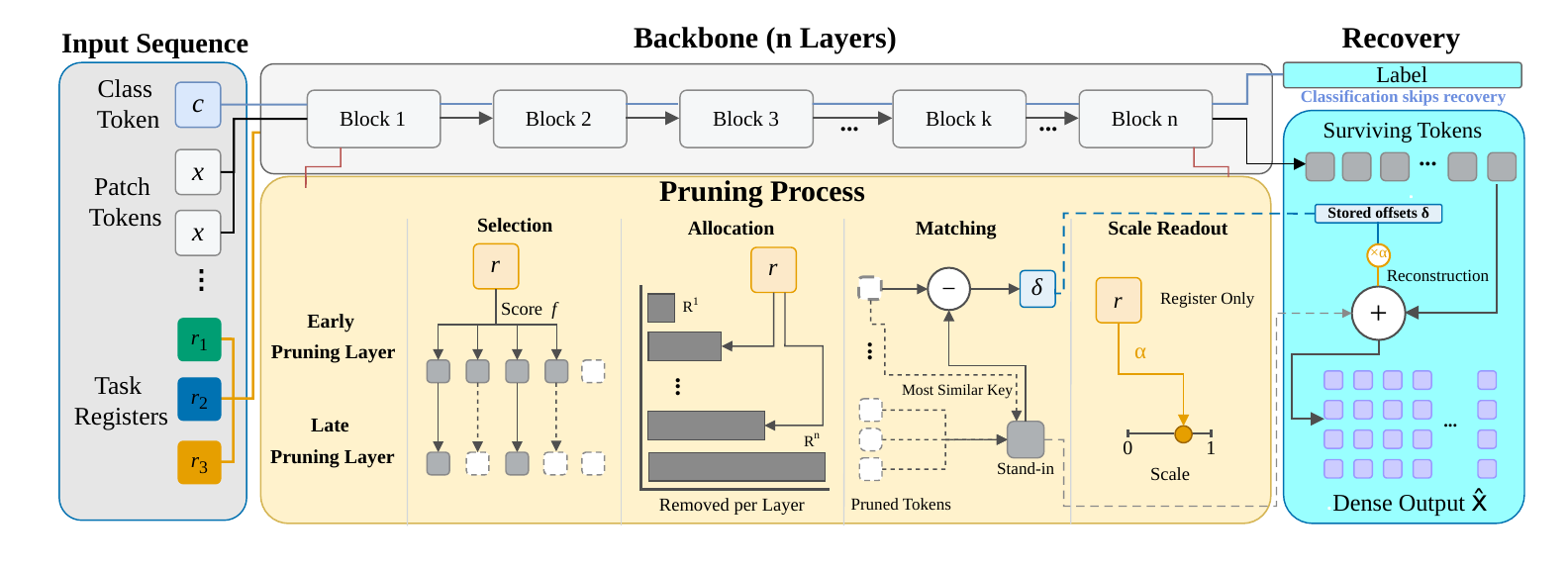}
\caption{Overview of TAP. Each task has its own learned initial register, and only the register associated with the current task is active in a forward pass. The active register evolves through the backbone and conditions token selection, exact budget allocation, and recovery scaling. Removed patches are matched to retained stand-ins, and reconstructed features are sent only to the dense readout.}
\label{fig:method}
\end{figure*}
Figure~\ref{fig:method} summarizes TAP, which maintains one learned initial register per task and propagates only the register associated with the current task. The active register evolves with the image features and conditions token selection, exact budget allocation, and recovery scaling. Each removed token is deterministically matched to the retained token with the highest key-space cosine similarity, and the register controls how much of the stored offset is restored. Token features remain image-dependent and can become task-specific through backbone fine-tuning in TAP-J or low-rank updates in TAP-F.

\subsubsection{Notation.}
A Vision Transformer splits an image into patch tokens and may prepend a class token $\mathbf{c}$. Let $\mathbf{X}^{(l)}=[\mathbf{x}^{(l)}_1,\dots,\mathbf{x}^{(l)}_{N_l}]$ denote the patch tokens entering layer $l$, where $\mathbf{x}^{(l)}_i\in\mathbb{R}^{d}$ and $N_l$ is the number of patch tokens entering that layer. The model processes one task $t\in\mathcal{T}$ at a time and prunes at $M$ ordered layers, $\mathcal{L}=\{l_1<\cdots<l_M\}$, where $M=|\mathcal{L}|$ is the number of pruning layers. We use $\sigma$ and $\mathrm{LN}$ to denote the logistic sigmoid and layer normalization, respectively. For compactness, the equations include the class token $\mathbf{c}$, which is omitted for tasks without one. At each pruning layer, tokens are scored and hard-selected before self-attention, so only the retained patch tokens are passed into the block.

\subsubsection{Task register.}
For each task $t\in\mathcal{T}$, TAP adds a learned register $\mathbf{r}^{(0)}_t\in\mathbb{R}^{d}$ whose dimension matches that of the patch tokens and class token. Standard transformer blocks update the register together with the image tokens. At a pruning layer, the block receives only the retained patch tokens together with the class token and register. Neither the class token nor the register is pruned. The evolving register state drives the decisions described below.

\subsubsection{Task-conditioned selection.}
At a pruning layer $l\in\mathcal{L}$, TAP normalizes the pre-attention states and reuses the block's query and key projections to compute a pre-softmax register-to-patch score. We define this score as
\begin{equation}
\label{eq:selection}
f^{(l)}_i=\sum_{h=1}^{H}
\frac{\big(\mathbf{W}^{(l,h)}_q\,\bar{\mathbf{r}}^{(l)}_t\big)^{\!\top}
\big(\mathbf{W}^{(l,h)}_k\,\bar{\mathbf{x}}^{(l)}_i\big)}
{\sqrt{d_h}},
\end{equation}
where $\mathbf{W}^{(l,h)}_q$ and $\mathbf{W}^{(l,h)}_k$ are the query and key projections for head $h$, and $d_h=d/H$ is the per-head dimension for $H$ heads. The normalized states are $\bar{\mathbf{r}}^{(l)}_t=\mathrm{LN}(\mathbf{r}^{(l)}_t)$ and $\bar{\mathbf{x}}^{(l)}_i=\mathrm{LN}(\mathbf{x}^{(l)}_i)$. For each head, scoring forms one register query and one key for each candidate patch by reusing the block's existing projections, so it introduces no additional projection parameters. At inference, TAP ranks the patch tokens by decreasing $f_i^{(l)}$ and resolves ties by ascending original patch index. During training, the hard forward pass applies the same ranking rule to the perturbed scores $z_i^{(l)}$ defined below. The required number of highest-ranked patches forms $\mathcal{S}^{(l)}$, while the remaining patches form $\mathcal{P}^{(l)}$. Only the patches in $\mathcal{S}^{(l)}$ enter full self-attention and the feed-forward sublayer together with the class token and task register, neither of which is ranked or pruned.

\subsubsection{Budget allocation.}
A learned allocation must satisfy the requested global token budget exactly. Let $N_1$ denote the initial number of patch tokens. A target keep rate $\rho\in(0,1]$ sets the final integer keep count $K$ and the removal budget $R=N_1-K$. At pruning layer $l$, $B^{(l)}$ denotes the unspent budget, $l^+$ denotes the next pruning layer, and $C^{(l)}=\min\{B^{(l)},N_l-1\}$ is the removal capacity. The sigmoid of a shared linear readout $(\mathbf{w},b)$ gives the fraction of $B^{(l)}$ proposed for that layer. Algorithm~\ref{alg:allocation} gives the resulting integer allocation.

Here $\widetilde R^{(l)}$ is the continuous proposal and $R^{(l)}$ is the integer removal count. The parameters $\mathbf{w}\in\mathbb{R}^{d}$ and $b\in\mathbb{R}$ are shared across tasks and layers. When no budget remains, $R^{(l)}=0$ and selection is skipped. After the preceding removals, $B^{(l)}=N_l-K$; hence the final budget cannot exceed $N_{l_M}-1$. The last layer can therefore consume the remainder while preserving at least one patch, which gives
$\sum_{l\in\mathcal{L}}R^{(l)}=R$ exactly. Training and inference both use the current image's register state.

\begin{algorithm}[!ht]
\caption{Register-conditioned exact-budget allocation}
\label{alg:allocation}
{\small
\begin{algorithmic}
\STATE $K \leftarrow
\max\!\left(1,\operatorname{round}(\rho N_1)\right)$,
\quad $R \leftarrow N_1-K$
\STATE $B^{(l_1)} \leftarrow R$
\FOR{$l=l_1,\ldots,l_{M-1}$}
    \STATE $\widetilde R^{(l)}
    \leftarrow
    \sigma\!\left(\mathbf{w}^{\!\top}\mathbf{r}^{(l)}_t+b\right)
    B^{(l)}$
    \STATE $R^{(l)}
    \leftarrow
    \min\!\left\{
    \operatorname{round}\!\left(\widetilde R^{(l)}\right),
    C^{(l)}
    \right\}$
    \STATE $B^{(l^+)}
    \leftarrow B^{(l)}-R^{(l)}$
\ENDFOR
\STATE $R^{(l_M)} \leftarrow B^{(l_M)}$
\end{algorithmic}
}
\end{algorithm}
\FloatBarrier

\subsubsection{Dense-feature recovery.}
Pruning removes a token from the backbone, but a dense head still requires a feature at its spatial position. Let $\mathbf{k}^{(l)}_i=\mathbf{W}^{(l)}_k\bar{\mathbf{x}}^{(l)}_i$, and let $\widetilde{\mathbf{k}}^{(l)}_i=\mathbf{k}^{(l)}_i/\max(\|\mathbf{k}^{(l)}_i\|_2,\epsilon)$ denote its normalized form for a small $\epsilon>0$. For each removed token $i\in\mathcal{P}^{(l)}$, we select the retained stand-in with the largest key-space cosine similarity:
\begin{equation}
\pi(i)=\operatorname*{arg\,max}_{j\in\mathcal{S}^{(l)}}
\big(\widetilde{\mathbf{k}}^{(l)}_i\big)^{\!\top}
\widetilde{\mathbf{k}}^{(l)}_j,
\end{equation}
where $\mathbf{W}^{(l)}_k$ is the concatenated multi-head key projection already computed for selection. We store the pointer $\pi(i)$ and the offset $\boldsymbol{\delta}_i=\mathbf{x}^{(l)}_i-\mathbf{x}^{(l)}_{\pi(i)}$. The stand-in remains in the sequence and evolves through later blocks. At a dense read point $L$, the removed position is reconstructed as
\begin{equation}
\hat{\mathbf{x}}_i=\mathbf{x}^{(L)}_{\pi^\ast(i)}+\alpha^{(l)}_t\,\boldsymbol{\delta}_i,
\qquad
\alpha^{(l)}_t=\sigma\big(\mathbf{u}^{\!\top}\mathbf{r}^{(l)}_t+v\big),
\end{equation}
where $\pi^\ast(i)$ follows the pointer chain to a token that survives at the read point. Reconstruction combines the original offset $\boldsymbol{\delta}_i$ with the final surviving endpoint and omits offsets from intermediate stand-ins. Those offsets arise at different depths and are not directly additive without a transport map. The shared readout, parameterized by $\mathbf{u}\in\mathbb{R}^{d}$ and $v\in\mathbb{R}$, controls how much of the early offset is added to the later endpoint. Values near one recover the full stored offset, while values near zero use the endpoint alone \citep{tokencropr}. Both training and inference compute $\alpha^{(l)}_t$ from the current register state.

\subsubsection{Sparse backbone, dense readout.}
Every surviving token retains its original patch index. At a feature read point, the recovery map reconstructs all positions removed up to that depth and scatters the surviving and reconstructed features back to the original grid. Reconstructed features are consumed only by the task head and never reenter the backbone. Classification reads the class token and skips recovery. Detection prunes only at global-attention blocks. Each subsequent window block groups surviving tokens by their original window and applies variable-length attention to every nonempty group. The original two-dimensional indices determine relative-position bias. No per-window quota is imposed, so global top-$k$ removes exactly $R^{(l)}$ tokens. If $n_\omega$ tokens survive in window $\omega$, variable-length packing incurs $\sum_\omega O(n_\omega^2d)$ attention cost without computation on padding.

\subsubsection{Training.}
Hard top-$k$ selection blocks gradients to the scoring and allocation readouts. Let $Q^{(l)}=\min\{\widetilde R^{(l)},C^{(l)}\}$ for $l\ne l_M$ and $Q^{(l_M)}=B^{(l_M)}$ denote the feasible continuous removal target. We train TAP with a cardinality-constrained straight-through mask. Let $\tau>0$ be the temperature, and let $\gamma_i$ be independent standard Gumbel samples \citep{gumbel}. When $Q^{(l)}>0$, $\theta^{(l)}$ is the unique solution to
\begin{equation}
\begin{array}{r@{\;}c@{\;}l}
z_i^{(l)} &=& f_i^{(l)}+\gamma_i,\\[1pt]
\displaystyle\sum_{i=1}^{N_l}
\sigma\!\left(\frac{z_i^{(l)}-\theta^{(l)}}{\tau}\right)
&=& N_l-Q^{(l)},\\[3pt]
m^{(l)}_{i,\mathrm{soft}}&=&
\displaystyle\sigma\!\left(\frac{z^{(l)}_i-\theta^{(l)}}{\tau}\right).
\end{array}
\end{equation}
We solve this monotone scalar equation by bisection and differentiate the solution implicitly. The hard mask is zero for the $R^{(l)}$ lowest perturbed scores and one otherwise. Its removed set is empty when $R^{(l)}=0$. The straight-through mask is
\begin{equation}
m^{(l)}_i=\operatorname{stopgrad}\!\left(m^{(l)}_{i,\mathrm{hard}}-m^{(l)}_{i,\mathrm{soft}}\right)+m^{(l)}_{i,\mathrm{soft}},
\end{equation}
where $\operatorname{stopgrad}$ is the identity in the forward pass and has zero derivative. The resulting hard-sparse feature input is
\begin{equation}
\mathbf{X}^{(l)}_{\mathrm{ST}}=
\operatorname{Gather}_{\mathcal{S}^{(l)}}\!
\left(\left[m^{(l)}_i\mathbf{x}^{(l)}_i\right]_{i=1}^{N_l}\right).
\end{equation}
Here $\operatorname{Gather}_{\mathcal{S}^{(l)}}$ selects the rows indexed by $\mathcal{S}^{(l)}$ and preserves their original patch order. The forward value is exactly the retained sequence because $m^{(l)}_i=1$ on the gathered indices. During backpropagation, the selected index set remains fixed, while the feature multipliers follow the soft mask~\citep{ste}. The implicitly differentiated threshold couples all candidates, allowing a removed score to affect the masks of retained candidates. Dense-task losses also reach removed features through recovery. We anneal $\tau$ during training and omit Gumbel noise at inference. TAP-J trains the backbone, register, readouts, and task head separately for each task. TAP-F freezes the base and trains on all three tasks jointly. At each step, we sample one task uniformly and activate its register, task head, and rank-8 LoRA updates in the attention and MLP linear maps~\citep{lora}. The shared allocation readout receives gradients from every task batch, but the recovery readout is updated only by dense-task batches. The low-rank updates can be merged for single-task inference. Because the hard forward pass already enforces the target budget, no rate loss is needed.

\subsubsection{Parameter and memory cost.}
TAP-J adds one register and two readouts, totaling $3d+2=2{,}306$ pruning-specific parameters for ViT-B. Recovery stores one pointer and one $d$-dimensional offset per removed token, requiring $R(d\,b_s+b_p)$ bytes for scalar width $b_s$ and pointer width $b_p$. TAP-F stores one frozen backbone and adds approximately $1.2$M low-rank parameters and one register per task, excluding the existing task heads.

\section{Experiments}

\subsubsection{Setup.}
We evaluate ImageNet-1K classification \citep{imagenet}, ADE20K semantic segmentation \citep{ade20k}, and joint object detection and instance segmentation on COCO \citep{coco}. Unless stated otherwise, all models use a ViT-B backbone with masked autoencoder (MAE) pretraining \citep{mae}. Classification uses a linear head, segmentation uses a SegFormer MLP head \citep{segformer}, and detection uses ViTDet \citep{vitdet} with Cascade Mask R-CNN \citep{cascade}. Detection and instance segmentation are optimized within the same detector pipeline and share one task register. Classification and segmentation prune at three evenly spaced layers. Detection prunes tokens in global attention blocks, after which window attention processes the survivors in groups of varying size.

We report top-1 accuracy, mean intersection over union (mIoU), and COCO box and mask average precision (AP). Segmentation uses five seeds, while classification and detection use three seeds. We measure throughput in images per second on a single NVIDIA RTX PRO 4000 Blackwell under mixed precision. Unless a value is marked as full-model throughput, it refers to the encoder. All methods within a task use the same input resolution, batch size, and timing procedure. The final keep rate $\rho$ is the fraction of input patch tokens that reach the last block.

TAP-J jointly fine-tunes a separate backbone and register for each task. TAP-F freezes one shared base and trains task-specific registers, low-rank updates, and heads together with shared allocation and recovery readouts. We compare TAP-J with fully fine-tuned methods and report TAP-F as a shared-base operating point.

All methods start from the same pretrained weights and task head and are fine-tuned with their pruning mechanisms. TAP-F throughput uses unmerged low-rank updates. A dagger marks a dense baseline without a native readout. We give these baselines the same parameter-free reconstruction: the most similar key-space stand-in and the validation-selected endpoint, with the full offset for segmentation and the stand-in alone for detection. The readout has no trainable parameters and is fixed across selectors.

\subsubsection{The controlled study.}
For each task, the probe freezes the corresponding no-pruning backbone and head, prunes each eligible layer in turn at a fixed token count without retraining, and measures the change from its no-pruning result. The parameter-free criteria span attention, activation magnitude, spatial coverage, and pairwise feature affinities used for merging. Learned selectors appear only as trained baselines in the main comparisons. Each task retains its native data, resolution, head, and eligible pruning layers, so the probes compare complete pipelines rather than task semantics alone.

Attention is the incoming patch-attention probability averaged over heads and query tokens, so it does not require a class token. Coverage applies deterministic farthest point sampling to normalized patch centers, starting with the patch nearest the image center. For affinity-based merging, the selected representative is retained and its pair removed, preserving the same hard token count. The recovery probe applies coverage at a local keep rate of $0.5$ over the same layers. Paired evaluations share retained sets, stand-ins, and offsets and differ only in $\alpha$: full offset uses $\alpha=1$ and stand-in only uses $\alpha=0$. The supplement lists all criteria and values.

\begin{table}[tb]
\centering
{\small
\begin{tabular}{lccc}
\toprule
Method & Size/add. & mIoU ($\Delta$) & im/s ($\times$) \\
\midrule
No pruning & 87.5M & 47.2 & 216 (1.00$\times$) \\
Random & +0 & 44.6 ($-$2.6) & 287 (1.33$\times$) \\
\midrule
Attn Top-K$^\dagger$ & +0 & 45.1 ($-$2.1) & 283 (1.31$\times$) \\
ToMe & +0 & 46.0 ($-$1.2) & 260 (1.20$\times$) \\
DiffRate$^\dagger$ & $<\!+0.1$K & 46.1 ($-$1.1) & 271 (1.25$\times$) \\
ALGM & +0 & 46.6 ($-$0.6) & 276 (1.28$\times$) \\
DToP & +0.5M & 46.2 ($-$1.0) & 253 (1.17$\times$) \\
Token Cropr & +2.6M & 46.7 ($-$0.5) & 268 (1.24$\times$) \\
\midrule
Static rule$^\dagger$ & +0 & 45.7 ($-$1.5) & 279 (1.29$\times$) \\
TAP-F & +1.2M & 45.8 ($-$1.4) & 275 (1.27$\times$) \\
TAP-J (ours) & +2.3K & \textbf{47.0} ($-$0.2) & 280 (1.30$\times$) \\
\bottomrule
\end{tabular}
}
\caption{Semantic segmentation on ADE20K at $\rho=0.5$. Parentheses report changes from no pruning, and throughput multipliers use the same reference. Size/add-on lists the no-pruning model once and then each method's addition. Fully fine-tuned rows require one backbone per task. TAP-F stores one frozen base and the listed task-specific add-on. The best fully fine-tuned result is bold.}
\label{tab:seg}
\end{table}
\begin{table}[tb]
\centering
{\small
\begin{tabular}{lccc}
\toprule
Method & Size/add. & Top-1 ($\Delta$) & im/s ($\times$) \\
\midrule
No pruning & 86.6M & 83.6 & 1830 (1.00$\times$) \\
Random & +0 & 82.0 ($-$1.6) & 2339 (1.28$\times$) \\
\midrule
Attn Top-K & +0 & 83.0 ($-$0.6) & 2325 (1.27$\times$) \\
EViT & +0 & 83.1 ($-$0.5) & 2277 (1.24$\times$) \\
DynamicViT & +2.0M & 83.0 ($-$0.6) & 2176 (1.19$\times$) \\
Evo-ViT & +0 & 82.6 ($-$1.0) & 1964 (1.07$\times$) \\
ATS & +0 & 82.9 ($-$0.7) & 2061 (1.13$\times$) \\
ToMe & +0 & 83.2 ($-$0.4) & 2202 (1.20$\times$) \\
DiffRate & $<\!+0.1$K & \textbf{83.3} ($-$0.3) & 2264 (1.24$\times$) \\
Token Cropr & +2.6M & 83.2 ($-$0.4) & 2225 (1.22$\times$) \\
\midrule
Static rule & +0 & 83.1 ($-$0.5) & 2314 (1.26$\times$) \\
TAP-F & +1.2M & 82.6 ($-$1.0) & 2288 (1.25$\times$) \\
TAP-J (ours) & +2.3K & 83.2 ($-$0.4) & 2300 (1.26$\times$) \\
\bottomrule
\end{tabular}
}
\caption{Image classification on ImageNet-1K at $\rho=0.5$. Layout, notation, and storage assumptions follow Table~\ref{tab:seg}.}
\label{tab:cls}
\end{table}
\begin{figure*}[t]
\centering
\includegraphics[width=0.86\textwidth]{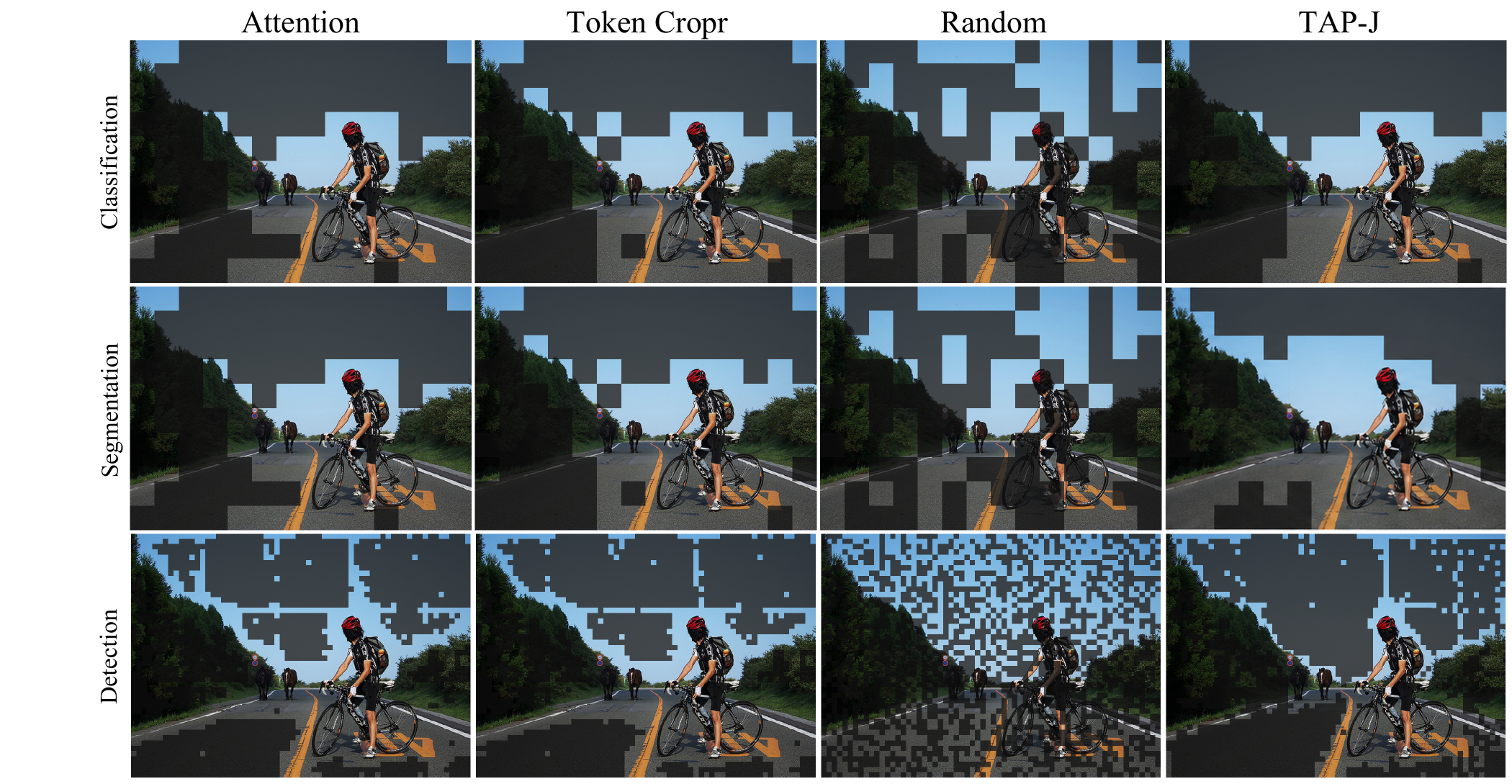}
\caption{Tokens that reach the last block for one image at $\rho=0.5$. Columns show attention, Token Cropr, random selection, and TAP-J, while rows show the three task settings. Removed tokens are darkened. Detection uses a higher input resolution and a finer token grid. In this example, TAP-J retains a spatially distributed set for segmentation and concentrates more tokens around objects for detection.}
\label{fig:qual}
\end{figure*}
\begin{table}[!t]
\centering
\setlength{\tabcolsep}{2.0pt}
{\small
\begin{tabular}{lcccc}
\toprule
Method & Size/add. & AP$^{\text{b}}$ & AP$^{\text{m}}$ & im/s (enc./full) \\
\midrule
No pruning & 132M & 54.0 & 46.7 & 14.2 / 9.6 \\
Random & +0 & 52.0 ($-$2.0) & 44.9 ($-$1.8) & 19.9 / 11.9 \\
\midrule
Attn Top-K$^\dagger$ & +0 & 53.1 ($-$0.9) & 45.9 ($-$0.8) & 19.7 / 11.8 \\
SViT & +0.6M & 53.5 ($-$0.5) & 46.2 ($-$0.5) & 17.9 / 11.2 \\
Token Cropr & +2.6M & 53.4 ($-$0.6) & 46.1 ($-$0.6) & 18.7 / 11.5 \\
\midrule
Static rule$^\dagger$ & +0 & 53.2 ($-$0.8) & 46.0 ($-$0.7) & 19.4 / 11.7 \\
TAP-F & +1.2M & 52.5 ($-$1.5) & 45.3 ($-$1.4) & 18.4 / 11.3 \\
TAP-J (ours) & +2.3K & \textbf{53.7} ($-$0.3) & \textbf{46.4} ($-$0.3) & 18.8 / 11.5 \\
\bottomrule
\end{tabular}
}
\caption{Object detection and instance segmentation on COCO at $\rho=0.5$. AP$^{\text{b}}$ and AP$^{\text{m}}$ denote box and mask AP, with changes from no pruning in parentheses. The final column reports encoder-only and full-model throughput in images per second. Size/add-on follows Table~\ref{tab:seg}, and the best fully fine-tuned results are bold.}
\label{tab:det}
\end{table}
\subsubsection{Main comparison.}
The static rule combines attention selection with one shared late-heavy schedule across tasks. TAP derives both choices from the current task register state.

Across Tables~\ref{tab:seg}--\ref{tab:det}, TAP-J has the smallest accuracy drop on segmentation and detection among the fully fine-tuned methods. On classification, its $83.2$ top-1 accuracy matches ToMe~\citep{tome} and Token Cropr~\citep{tokencropr}, and trails DiffRate~\citep{diffrate} by $0.1$ point. Compared with the static rule, TAP-J gains $0.1$ point on classification, $1.3$ mIoU on segmentation, and $0.5$ box AP on detection. The largest gains occur on the dense tasks, linking the conflicts in criterion ranking, pruning depth, and recovery identified in Figure~\ref{fig:insight}(b--d) to TAP's task-conditioned decisions. TAP-F reduces task-specific storage by sharing the frozen base. With approximately $1.2$M task-specific parameters, it trails no pruning by $1.0$ top-1 points, $1.4$ mIoU, and $1.5$ box AP.

Figure~\ref{fig:qual} illustrates task-dependent selection, complementing the dataset-level evidence in Figure~\ref{fig:insight}.

\subsubsection{Ablation.}
\begin{table}[!t]
\centering
\setlength{\tabcolsep}{4.2pt}
{\small
\begin{tabular}{lccc}
\toprule
Variant & cls & seg & det \\
\midrule
TAP-F (full) & 0.0 & 0.0 & 0.0 \\
\midrule
attention selection & $-$0.2 & $-$0.9 & $-$0.5 \\
static task query & $-$0.1 & $-$0.4 & $-$0.3 \\
\midrule
recovery off (zero-fill) & n/a & $-$3.4 & $-$2.2 \\
$\alpha\equiv 1$ (full offset) & n/a & $-$0.1 & $-$1.0 \\
$\alpha\equiv 0$ (stand-in copy) & n/a & $-$1.6 & $-$0.1 \\
\midrule
task adapters, shared register & $-$0.2 & $-$1.3 & $-$1.1 \\
\midrule
task-shared allocation & $-$0.1 & $-$0.6 & $-$0.5 \\
layer-uniform allocation & $-$0.6 & $-$1.2 & $-$0.8 \\
task-mean allocation & $-$0.1 & $-$0.2 & $-$0.1 \\
task-mean recovery & n/a & $-$0.1 & $-$0.1 \\
both task means & $-$0.1 & $-$0.3 & $-$0.2 \\
\bottomrule
\end{tabular}
}
\caption{Selection, recovery, register, and allocation ablations at $\rho=0.5$ in the frozen-base regime. Results are changes from TAP-F. The notation n/a marks recovery settings that do not apply to classification. Cls, Seg, and Det denote classification, segmentation, and detection.}
\label{tab:ablation}
\end{table}
Table~\ref{tab:ablation} directly tests the selection, depth, and recovery conflicts identified in Figure~\ref{fig:insight} by ablating TAP's three register-driven choices. Replacing Equation~\eqref{eq:selection} with incoming patch attention loses $0.9$ mIoU and $0.5$ box AP. Using the initial register instead of its evolving state costs $0.4$ mIoU and $0.3$ box AP. Zero-filling is worse than both recovery endpoints, and the preferred endpoint differs between segmentation and detection. Sharing the initial register across task-specific adapters costs $0.2$ top-1 points, $1.3$ mIoU, and $1.1$ box AP, although the shared state still receives task-specific adapter features. Task-shared and layer-uniform controls show that allocation depends on task and depth. Replacing per-image allocation and recovery with layerwise validation means costs $0.3$ mIoU and $0.2$ box AP.

\begin{figure}[!tb]
\centering
\includegraphics[width=\columnwidth]{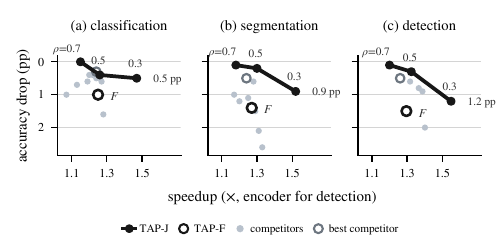}
\caption{Accuracy drop against encoder speedup with the vertical axis inverted. Competitors use $\rho=0.5$, TAP-J uses $\rho\in\{0.3,0.5,0.7\}$, and the open symbol marks TAP-F. Outlined symbols identify the strongest competitor for each task.}
\label{fig:tradeoff}
\end{figure}
\subsubsection{Efficiency and stability.}\label{sec:efficiency}
Figure~\ref{fig:tradeoff} summarizes TAP-J's accuracy-throughput trade-off across all three tasks. At $\rho=0.5$, TAP-J reaches encoder speedups of $1.26\times$, $1.30\times$, and $1.32\times$ with corresponding losses of $0.4$ top-1 points, $0.2$ mIoU, and $0.3$ box AP. Its dense-task losses are the smallest among the compared methods. The unchanged detection head limits the full-model speedup to $1.20\times$. The register query, budget readout, and top-$k$ bookkeeping account for $0.7\%$ of encoder runtime.

Throughput includes recovery matching, sparse packing, and reconstruction. Standard deviations are at most $0.24$, and Table~\ref{tab:stability-memory} shows lower peak inference memory for every task and regime. Component profiles are in the supplement.

\begin{figure}[!ht]
\centering
\includegraphics[width=\columnwidth]{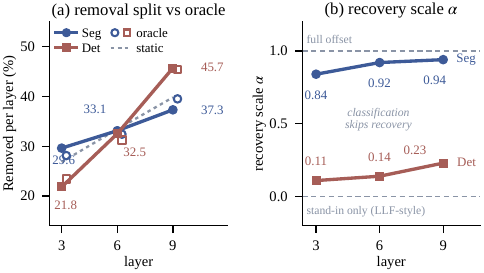}
\caption{Register-conditioned validation summaries. (a) Mean layerwise removal allocation with the oracle (open markers) and static schedule. (b) Mean recovery scale by dense task and layer.}
\label{fig:sandwich}
\end{figure}
\subsubsection{What does the register learn?}
Figure~\ref{fig:sandwich}(a) compares the learned allocation with a diagnostic oracle that exhaustively selects the best integer layer split on held-out data under the same budget. The oracle is used only for this analysis. TAP removes fewer tokens early and learns different mean schedules for segmentation and detection. Figure~\ref{fig:sandwich}(b) places segmentation near the full-offset endpoint and detection near the stand-in endpoint, consistent with the paired recovery probe. The curves report validation-set means, but allocation and recovery are computed per image from the register.

\begin{table}[!ht]
\centering
\setlength{\tabcolsep}{2.3pt}
{\small
\begin{tabular}{lccccc}
\toprule
Task & Score & Std. & Base & Peak GB & $\Delta$ (\%) \\
\midrule
Cls & 83.2/82.6 & 0.06/0.08 & 0.42 & 0.37/0.38 & $-$11.9/$-$9.5 \\
Seg & 47.0/45.8 & 0.18/0.24 & 0.78 & 0.64/0.66 & $-$17.9/$-$15.4 \\
Det & 53.7/52.5 & 0.11/0.14 & 3.10 & 2.82/2.88 & $-$9.0/$-$7.1 \\
\bottomrule
\end{tabular}
}
\caption{Run-to-run variation and peak inference memory at $\rho=0.5$. Slashes report TAP-J/TAP-F. Score denotes top-1 accuracy, mIoU, or box AP for Cls, Seg, or Det. Base is the no-pruning peak.}
\label{tab:stability-memory}
\end{table}

\FloatBarrier
\section{Discussion}
The evidence argues against treating token reduction as a single transferable rule. As ViT representations move from local structure to higher-level semantics, token value changes with depth and with the task that reads the backbone. TAP uses the evolving register to turn this internal state into task-dependent computational decisions.

TAP-J trails DiffRate by $0.1$ point on classification. Classification reads a class token and skips dense recovery, leaving one of TAP's three mechanisms unused. Its global objective also supplies weaker spatial guidance than segmentation masks or detection boxes. These factors leave a classification-specific rate schedule competitive.

These results point to finer-grained task-adaptive computation. Segmentation may need broad spatial support across more layers, while detection may concentrate computation once object-level evidence emerges. Depth estimation could preserve geometry, video could retain temporally stable regions, and language-conditioned vision could focus on prompt-relevant content. Layerwise policies may capture these transitions and coordinate pruning with intermediate feature reads. The present design fixes the total keep rate, requires task-specific training, and reconstructs each removed position from one stand-in and one earlier offset. Future work should examine image-conditioned budgets, multiple stand-ins, cross-layer transport, and matched pipelines that separate task semantics from implementation differences. As vision backbones grow, our longer-term goal is a unified model that adapts sparse computation across more visual tasks with minimal task-specific storage.

\section{Conclusion}
Our preliminary experiments show that pruning behavior changes with task and depth, which motivates us to propose Task-Adaptive Pruning (TAP). Rather than relying on a pruning rule tailored to one visual task, TAP puts an evolving task register to work as the controller of a shared sparse computation process under an exact global budget. This simple and efficient design maintains a favorable balance between predictive performance and encoder throughput across tasks, making TAP particularly well suited to unified vision systems that serve multiple tasks through a shared backbone.

\bibliography{aaai2027}

\end{document}


\maketitle

\section*{Supplementary Overview}

This document supplements ``Putting Registers to Work: Task Registers for Token Pruning in Vision Transformers.'' It expands the method, experiments, and reproducibility record of Task-Adaptive Pruning (TAP). Section~\ref{sec:complete-method} gives the computation order and proves the exact-budget property. Section~\ref{sec:implementation} records the shared implementation and task-specific settings. Section~\ref{sec:probes} defines the controlled probes. Sections~\ref{sec:extended-results}--\ref{sec:register-analysis} extend the main comparisons, ablations, and register analysis. Section~\ref{sec:unified-backbone} examines what a unified ViT backbone can share without imposing one pruning policy on every task. Section~\ref{sec:efficiency} separates policy overhead from sparse execution and dense reconstruction. The final sections cover spatial retention, limitations, and the claim-to-evidence map.

\section{TAP Formulation and Algorithms}
\label{sec:complete-method}

\subsection{Notation and computation order}

TAP processes one task $t\in\mathcal{T}$ during a forward pass. The Vision Transformer has $L$ blocks and prunes at the ordered subset $\mathcal{L}=\{l_1<\cdots<l_M\}$. The patch sequence entering block $l$ is $\mathbf{X}^{(l)}=[\mathbf{x}^{(l)}_1,\ldots,\mathbf{x}^{(l)}_{N_l}]$, where each token has dimension $d$. Each token retains its original spatial index after earlier tokens have been removed. A class token is present when the task pipeline uses one. Every task has a learned initial register $\mathbf{r}^{(0)}_t\in\mathbb{R}^d$, and only the register for the active task enters the backbone.

Table~\ref{tab:notation} collects the symbols used below. Batch dimensions are omitted because selection, allocation, and recovery are computed independently for each image. The global keep rate fixes the final integer token count. The register may change the distribution of removals across depth, but it cannot change that final count.

Three constraints define the computation. The task register remains in the sparse sequence and is never a pruning candidate. A removed patch cannot re-enter the encoder, even when a dense head later requests its feature. The final keep count is fixed before the first pruning decision. The register controls only how that budget is distributed. These constraints keep recovery outside the encoder and preserve the stated sparse workload.

\begin{table*}[t]
\centering
\begin{tabular}{lll}
\toprule
Symbol & Meaning & Shape or domain \\
\midrule
$t,\mathcal{T}$ & active task and task set & $t\in\mathcal{T}$ \\
$L,\mathcal{L},M$ & backbone depth, pruning layers, number of pruning layers & $\mathcal{L}=\{l_1<\cdots<l_M\}$ \\
$\mathbf{X}^{(l)},N_l$ & patch sequence and its length before block $l$ & $\mathbf{X}^{(l)}\in\mathbb{R}^{N_l\times d}$ \\
$\mathbf{r}^{(l)}_t$ & state of the active task register & $\mathbb{R}^{d}$ \\
$f_i^{(l)}$ & register-to-patch selection score & $\mathbb{R}$ \\
$\rho,K,R$ & keep rate, final keep count, total removal budget & $\rho\in(0,1]$, $K\in\mathbb{N}$, $R\in\mathbb{N}_0$ \\
$B^{(l)},C^{(l)}$ & remaining budget and current removal capacity & $\mathbb{N}_0$ \\
$\widetilde R^{(l)},R^{(l)}$ & continuous proposal and integer removal count & $\mathbb{R}_{\geq0}$, $\mathbb{N}_0$ \\
$Q^{(l)}$ & feasible continuous removal target used in training & $[0,N_l-1]$ \\
$\mathcal{S}^{(l)},\mathcal{P}^{(l)}$ & retained and removed patch sets & subsets of current indices \\
$\pi(i),\pi^\ast(i)$ & immediate stand-in and final surviving endpoint & token indices \\
$\boldsymbol{\delta}_i$ & feature offset stored at the removal layer & $\mathbb{R}^{d}$ \\
$\alpha_t^{(l)}$ & register-conditioned recovery scale & $(0,1)$ \\
$m_{i,\mathrm{hard}}^{(l)},s_i^{(l)}$ & hard keep decision and soft keep probability & $\{0,1\}$, $(0,1)$ \\
\bottomrule
\end{tabular}
\caption{Notation used in the supplementary derivations. The original spatial index is carried separately from the row position in the current sparse sequence.}
\label{tab:notation}
\end{table*}

At a pruning block, TAP first normalizes the current states and computes one register query together with one key for every candidate patch. The score reuses the block projections and is available before full self-attention. The allocator then determines the number of patches removed at this layer. The selected patches, the class token if present, and the task register enter the block. Removed patches do not return to later backbone blocks. Dense features are reconstructed only when a downstream head requests a complete spatial grid.

\begin{algorithm}[t]
\caption{TAP forward pass for task $t$}
\label{alg:tap-forward}
\begin{algorithmic}[1]
\REQUIRE Patch tokens $\mathbf{X}$ with original indices, active register $\mathbf{r}_t^{(0)}$, task head, keep rate $\rho$, pruning layers $\mathcal{L}$
\STATE Set $K\leftarrow\max\{1,\operatorname{round}(\rho N_1)\}$, $R\leftarrow N_1-K$, and $B\leftarrow R$
\STATE Initialize an empty dense-feature collection $\mathcal{D}$
\FOR{$l=1,\ldots,L$}
  \IF{$l\in\mathcal{L}$ and $B>0$}
    \STATE Compute register-to-patch scores from the pre-attention normalized states
    \STATE Convert the current register state and remaining budget into $R^{(l)}$
    \STATE Rank patches by score and original index, then form $\mathcal{S}^{(l)}$ and $\mathcal{P}^{(l)}$
    \IF{task $t$ uses a dense readout}
      \FOR{each $i\in\mathcal{P}^{(l)}$}
        \STATE Match $i$ to its most similar retained key $\pi(i)$ and store $\boldsymbol{\delta}_i$
      \ENDFOR
      \STATE Compute and store the scalar recovery scale $\alpha_t^{(l)}$
    \ENDIF
    \STATE Gather the retained sequence and update $B\leftarrow B-R^{(l)}$
  \ENDIF
  \STATE Apply the transformer block to the current sparse sequence
  \IF{the task head reads a dense feature grid after block $l$}
    \STATE Follow each stored pointer to its surviving endpoint
    \STATE Reconstruct removed positions and scatter all positions to the original grid
    \STATE Append the dense grid to $\mathcal{D}$ without reinserting it into the backbone
  \ENDIF
\ENDFOR
\RETURN The task head applied to the final class token or the collected grids $\mathcal{D}$
\end{algorithmic}
\end{algorithm}

\subsection{Task-conditioned selection}

Let $H$ be the number of attention heads and $d_h=d/H$ the dimension of each head. We write $\bar{\mathbf{r}}_t^{(l)}=\operatorname{LN}(\mathbf{r}_t^{(l)})$ and $\bar{\mathbf{x}}_i^{(l)}=\operatorname{LN}(\mathbf{x}_i^{(l)})$. For head $h$, TAP applies the existing query projection to the normalized register and the existing key projection to every normalized patch. Summing the pre-softmax logits over heads gives
\begin{equation}
f_i^{(l)}=\sum_{h=1}^{H}
\frac{\big(\mathbf{W}^{(l,h)}_q\bar{\mathbf{r}}^{(l)}_t\big)^\top
\big(\mathbf{W}^{(l,h)}_k\bar{\mathbf{x}}^{(l)}_i\big)}{\sqrt{d_h}}.
\label{eq:supp-selection}
\end{equation}
The sum and mean over heads induce the same hard ranking because $H$ is fixed within a block. Training uses the sum throughout, and the reported temperature and Gumbel scale follow this convention. The class token and task register bypass the ranking. TAP orders patch tokens by decreasing score and uses the ascending original patch index to resolve an exact tie. The resulting selection is deterministic at inference and preserves the requested cardinality.

At a pruning block, the implementation slices the fused query-key-value projection. It first computes the key projection for every candidate patch and the query projection for the active register. After hard selection, it gathers the retained keys and computes the remaining query and value projections only for the sequence that enters attention. The class token and register receive any projections not already computed. Non-pruning blocks retain the original fused projection path. This ordering reuses the candidate keys for matching and avoids computing patch queries, values, attention, or MLP activations for removed tokens.

Algorithm~\ref{alg:selection} makes the index handling explicit. Each retained token keeps its original spatial index, which recovery and relative-position bias later reuse.

\begin{algorithm}[t]
\caption{Hard selection with stable spatial indices}
\label{alg:selection}
\begin{algorithmic}[1]
\REQUIRE Ranking scores $q_i^{(l)}$, original indices $o_i$, integer removal count $R^{(l)}$
\IF{$R^{(l)}=0$}
  \RETURN all patches in their current order
\ENDIF
\STATE Sort candidate patches by decreasing $q_i^{(l)}$ and then by increasing $o_i$
\STATE Retain the first $N_l-R^{(l)}$ candidates and remove the remainder
\STATE Restore the retained candidates to their original sequence order before gathering
\RETURN retained and removed sets with their original indices
\end{algorithmic}
\end{algorithm}

During training, Algorithm~\ref{alg:selection} uses $q_i^{(l)}=z_i^{(l)}$. Inference sets $q_i^{(l)}=f_i^{(l)}$ and introduces no perturbation.

\subsection{Exact global budget}

The target keep rate defines $K=\max\{1,\operatorname{round}(\rho N_1)\}$ and $R=N_1-K$. We implement rounding as $\operatorname{round}(x)=\lfloor x+1/2\rfloor$ for nonnegative $x$, which removes any dependence on a library's tie convention. Before any removal, $B^{(l_1)}=R$. At a non-final pruning layer, the readout $\mathbf{w}\in\mathbb{R}^d$ and $b\in\mathbb{R}$ proposes a fraction of the unspent budget. Capacity clipping prevents that layer from removing every remaining patch:
\begin{equation}
\begin{aligned}
g^{(l)} &= \mathbf{w}^{\top}\mathbf{r}_t^{(l)}+b, \\
\widetilde R^{(l)} &= \sigma(g^{(l)})B^{(l)}, \\
C^{(l)} &= \min\{B^{(l)},N_l-1\}, \\
R^{(l)} &= \min\{\operatorname{round}(\widetilde R^{(l)}),C^{(l)}\}, \quad l\ne l_M, \\
B^{(l^+)} &= B^{(l)}-R^{(l)}.
\end{aligned}
\label{eq:supp-budget}
\end{equation}
The final pruning layer sets $R^{(l_M)}=B^{(l_M)}$. This assignment removes the rounding residue and produces the exact global count.

\begin{proposition}[Exact budget and nonempty output]
Assume that pruning is the only operation that changes the number of patch tokens. The recursion in Equation~\ref{eq:supp-budget} removes exactly $R=N_1-K$ patches and leaves exactly $K\geq1$ patches after the final pruning layer.
\end{proposition}

\noindent\textit{Proof.}
Let $A_l$ denote the number of patches removed before pruning layer $l$. The current sequence has $N_l=N_1-A_l$ patches, and the remaining budget is
\begin{equation}
B^{(l)}=R-A_l=N_1-K-A_l=N_l-K.
\label{eq:budget-invariant}
\end{equation}
The equality holds at $l_1$ because $A_{l_1}=0$. If a non-final layer removes $R^{(l)}$, both $N_l$ and $B^{(l)}$ decrease by the same integer, so the invariant is preserved at the next pruning layer. Capacity clipping gives $R^{(l)}\leq N_l-1$ and leaves at least one patch after every non-final removal. At $l_M$, Equation~\ref{eq:budget-invariant} gives $B^{(l_M)}=N_{l_M}-K\leq N_{l_M}-1$ because $K\geq1$. The final layer can remove the entire remainder without emptying the sequence. The total removal is $R-B^{(l_M)}+B^{(l_M)}=R$, and the final sequence contains $N_1-R=K$ patches. \hfill$\square$

Table~\ref{tab:budget-tests} illustrates the invariant with synthetic unit tests. These cases test the integer logic and do not represent empirical results.

\begin{table}[b]
\centering
\begin{tabular}{rrrrl}
\toprule
$N_1$ & $K$ & proposals & removals & check \\
\midrule
196 & 196 & 0, 0, 0 & 0, 0, 0 & keep all \\
196 & 137 & 18, 20, rem. & 18, 20, 21 & sum $=59$ \\
196 & 98 & 25, 31, rem. & 25, 31, 42 & sum $=98$ \\
196 & 59 & 0, 68, rem. & 0, 68, 69 & sum $=137$ \\
4 & 1 & 3, rem. & 3, 0 & one survives \\
\bottomrule
\end{tabular}
\caption{Synthetic checks of exact integer allocation. ``rem.'' assigns the remaining budget to the final pruning layer.}
\label{tab:budget-tests}
\end{table}

\subsection{Cardinality-constrained straight-through training}

The hard ranking used by the sparse forward pass is not differentiable. TAP retains the hard sequence length and supplies a soft gradient through a constrained mask. The feasible continuous target is $Q^{(l)}=\min\{\widetilde R^{(l)},C^{(l)}\}$ at a non-final pruning layer and $Q^{(l_M)}=B^{(l_M)}$ at the final one. During training, perturbed scores are $z_i^{(l)}=f_i^{(l)}+\gamma_i$, where $\gamma_i$ follows a standard Gumbel distribution~\citep{gumbel}. The soft keep probabilities satisfy
\begin{align}
s_i &= \sigma\!\left(\frac{z_i-\theta}{\tau}\right), \\
\sum_{i=1}^{N_l}s_i &= N_l-Q^{(l)}.
\label{eq:soft-cardinality}
\end{align}
For $0<Q^{(l)}<N_l$, the left side is continuous and strictly decreasing in $\theta$. Its limits are $N_l$ and zero, so Equation~\ref{eq:soft-cardinality} has a unique solution. We solve for $\theta$ in FP32 with 32 bisection steps. The initial interval is $[\min_i z_i-20\tau,\max_i z_i+20\tau]$. For the sequence lengths used here, this bracket places the endpoint residual below $10^{-5}$. When $Q^{(l)}=0$, TAP keeps every token and skips the solve.

The implicit derivative also has a closed form. Let $a_i=s_i(1-s_i)$ and $A=\sum_i a_i$. Holding $Q^{(l)}$ fixed gives
\begin{equation}
\frac{\partial\theta}{\partial z_j}=\frac{a_j}{A}, \qquad
\frac{\partial s_i}{\partial z_j}=\frac{a_i}{\tau}
\left(\mathbb{I}[i=j]-\frac{a_j}{A}\right).
\label{eq:implicit-score}
\end{equation}
When the allocation readout changes $Q^{(l)}$, the additional derivative is
\begin{equation}
\frac{\partial\theta}{\partial Q^{(l)}}=\frac{\tau}{A}, \qquad
\frac{\partial s_i}{\partial Q^{(l)}}=-\frac{a_i}{A}.
\label{eq:implicit-budget}
\end{equation}
These expressions are exact for $A>0$. In the implementation, every denominator $A$ is replaced by $A_\epsilon=\max(A,10^{-6})$ to avoid numerical amplification after sigmoid saturation. Gradients pass through $Q^{(l)}$ while $\widetilde R^{(l)}<C^{(l)}$ and stop at the capacity bound, with the zero subgradient used at equality. The integer budget passed to the next layer is detached, so each allocation receives a local gradient and the final layer consumes the remaining hard budget. The negative sign in Equation~\ref{eq:implicit-budget} reflects the cardinality constraint: a larger removal target lowers the total soft keep mass.

The hard mask keeps the $N_l-R^{(l)}$ highest perturbed scores. The straight-through mask is
\begin{equation}
m_i=\operatorname{stopgrad}(m_{i,\mathrm{hard}}-s_i)+s_i.
\end{equation}
Its forward value is binary, while its derivative follows the constrained soft mask~\citep{ste}. Gathering the hard retained set after multiplying by $m_i$ keeps the actual encoder sparse. It also allows the scoring and allocation readouts to receive gradients through the retained features. Dense-task losses provide a second route through reconstructed removed features. The temperature follows a cosine schedule from $1.0$ to $0.1$. Gumbel perturbations are sampled only during training. Inference ranks the unperturbed scores and is deterministic. The exact hard count removes the need for an auxiliary rate loss.

\begin{figure}[t]
\centering
\begin{tikzpicture}[
font=\scriptsize,
box/.style={draw, rounded corners=1pt, align=center, minimum height=5mm, inner sep=2pt},
flow/.style={-{Latex[length=1.5mm]}, line width=0.45pt},
grad/.style={-{Latex[length=1.5mm]}, dashed, line width=0.45pt},
node distance=4mm and 4mm]
\node[box] (score) {scores\\$f^{(l)}$};
\node[box, right=of score] (perturb) {training scores\\$z^{(l)}$};
\node[box, above right=3mm and 5mm of perturb] (hard) {hard top-$k$\\$m_{\rm hard}$};
\node[box, below right=3mm and 5mm of perturb] (soft) {cardinality solve\\$s,\theta$};
\node[box, right=11mm of perturb] (st) {straight-through\\mask $m$};
\node[box, right=of st] (encoder) {sparse\\encoder};
\draw[flow] (score) -- (perturb);
\draw[flow] (perturb) -- (hard);
\draw[flow] (perturb) -- (soft);
\draw[flow] (hard) -- (st);
\draw[grad] (soft) -- (st);
\draw[flow] (st) -- (encoder);
\node[above=1mm of hard] {binary forward};
\node[below=1mm of soft] {constrained gradient};
\end{tikzpicture}
\caption{Training computation at one pruning layer. The encoder receives the binary hard selection. The backward path follows a soft mask whose total keep mass matches the continuous target.}
\label{fig:training-graph}
\end{figure}
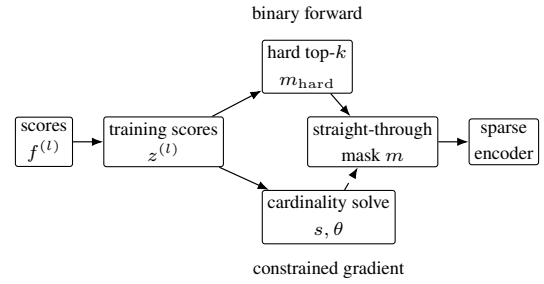

\subsection{Stand-in matching and dense recovery}

For each removed token, TAP selects a retained stand-in in the normalized concatenated multi-head key space. Let $\mathbf{k}_i^{(l)}$ denote the concatenation of the keys already computed for selection and let $\widetilde{\mathbf{k}}_i^{(l)}=\mathbf{k}_i^{(l)}/\max\{\|\mathbf{k}_i^{(l)}\|_2,\epsilon\}$ for a small $\epsilon>0$. The immediate pointer and offset are
\begin{align}
\pi(i)&=\operatorname*{arg\,max}_{j\in\mathcal{S}^{(l)}}
\widetilde{\mathbf{k}}_i^{(l)\top}\widetilde{\mathbf{k}}_j^{(l)}, \\
\boldsymbol{\delta}_i&=\mathbf{x}^{(l)}_i-\mathbf{x}^{(l)}_{\pi(i)}.
\end{align}
If an earlier stand-in is removed at a later layer, its pointer extends the chain to a token that remains active. At a read point $l_{\mathrm{read}}\geq l$, following that chain yields $\pi^\ast(i)$. TAP reconstructs the removed position as
\begin{equation}
\widehat{\mathbf{x}}_i=\mathbf{x}^{(l_{\mathrm{read}})}_{\pi^\ast(i)}+
\alpha_t^{(l)}\boldsymbol{\delta}_i, \qquad
\alpha_t^{(l)}=\sigma(\mathbf{u}^{\top}\mathbf{r}^{(l)}_t+v).
\label{eq:supp-recovery}
\end{equation}
Here $\mathbf{u}\in\mathbb{R}^d$ and $v\in\mathbb{R}$ form the scalar recovery readout.
The endpoint carries the later semantic state of the surviving path. The stored offset preserves the difference observed when token $i$ left the encoder. All ViT blocks retain the same residual-stream dimension, so the early offset can serve as a correction in the shared channel coordinates at a later read point. This is an approximation, since the intervening blocks do not provide an explicit transport map. The learned scale limits the contribution of an offset when that approximation is unreliable. TAP does not add offsets created by intermediate stand-ins because they arise at different depths and are not directly additive.

Stand-in assignment is deterministic. An exact similarity tie is resolved by the smaller original patch index. The pointer and its index operations are treated as discrete and receive no gradient. Gradients still pass through the removed feature, the selected endpoint, the stored offset, and the recovery scale. The implementation computes $\alpha_t^{(l)}$ once at each pruning layer and stores the single value for all tokens removed there. It does not retain the full register state for recovery.

The implementation forms all stand-in similarities at a layer with one matrix multiplication between the normalized removed-key and retained-key matrices. Pointer chains contain at most $M-1$ links and are resolved with batched index gathers at a dense read point.

\begin{algorithm}[t]
\captionsetup{singlelinecheck=false,justification=raggedright}
\caption{Pointer-chain recovery}
\label{alg:recovery}
\begin{algorithmic}[1]
\REQUIRE Surviving features at read point $l_{\mathrm{read}}$, stored pointers, stored offsets, original spatial indices
\FOR{each position removed at layer $l\leq l_{\mathrm{read}}$}
  \STATE Follow its pointer chain until the endpoint survives at $l_{\mathrm{read}}$
  \STATE Read the scalar $\alpha_t^{(l)}$ saved when the token was removed
  \STATE Reconstruct with Equation~\ref{eq:supp-recovery}
\ENDFOR
\STATE Scatter surviving and reconstructed features to the original spatial grid
\RETURN Dense grid for the task head
\end{algorithmic}
\end{algorithm}

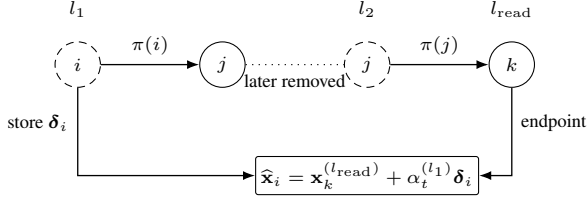
\begin{figure}[t]
\centering
\begin{tikzpicture}[
font=\scriptsize,
tok/.style={circle, draw, minimum size=6mm, inner sep=0pt},
removed/.style={tok, densely dashed},
ptr/.style={-{Latex[length=1.6mm]}, line width=0.5pt},
node distance=9mm and 13mm]
\node[removed] (i) {$i$};
\node[tok, right=of i] (j1) {$j$};
\node[removed, right=of j1] (j2) {$j$};
\node[tok, right=of j2] (k) {$k$};
\node[above=2mm of i] {$l_1$};
\node[above=2mm of j2] {$l_2$};
\node[above=2mm of k] {$l_{\mathrm{read}}$};
\draw[ptr] (i) -- node[above] {$\pi(i)$} (j1);
\draw[ptr] (j2) -- node[above] {$\pi(j)$} (k);
\draw[dotted, line width=0.45pt] (j1) -- node[below] {later removed} (j2);
\node[draw, rounded corners=1pt, below=9mm of j2, align=center, inner sep=2pt] (rec)
{$\widehat{\mathbf{x}}_i=\mathbf{x}^{(l_{\mathrm{read}})}_k+\alpha_t^{(l_1)}\boldsymbol{\delta}_i$};
\draw[ptr] (i) |- node[pos=0.2, left] {store $\boldsymbol{\delta}_i$} (rec);
\draw[ptr] (k) |- node[pos=0.2, right] {endpoint} (rec);
\end{tikzpicture}
\caption{Pointer resolution across pruning layers. If the first stand-in is later removed, its pointer extends the chain to survivor $k$. Recovery uses the final endpoint and the offset stored when $i$ was removed. Intermediate offsets are not accumulated.}
\label{fig:pointer-chain}
\end{figure}

\subsection{Sparse window attention for detection}

Detection follows the global and windowed attention structure of ViTDet~\citep{vitdet}. The task register joins the sequence and is updated at global-attention blocks. It bypasses each intervening window-attention block without being copied into the windows. Patch tokens continue to evolve inside their assigned windows, so the next global block scores current patch keys using the register state produced by the previous global interaction. The reported TAP models prune at blocks 3, 6, and 9, while the controlled probe also evaluates the final global-attention block. Each later window block assigns surviving tokens to windows using their original two-dimensional indices. For window $\omega$, let $\mathcal{I}_\omega$ contain the surviving tokens whose original positions lie in that window. Attention is evaluated only for nonempty groups, and the relative-position bias uses pairs of original positions. The attention cost is
\begin{equation}
\sum_{\omega:\,|\mathcal{I}_\omega|>0}O(|\mathcal{I}_\omega|^2d).
\end{equation}
TAP imposes no per-window quota, so the global ranking determines which tokens are removed and each image still removes exactly $R^{(l)}$ tokens. The implementation concatenates the nonempty windows, records their cumulative sequence lengths, and evaluates them with a variable-length attention kernel. Relative-position terms are gathered from the original two-dimensional indices before the kernel call. Packing, bias gathering, and scattering are included in the reported latency. This global-only register update avoids a separate rule for combining window-specific register copies and keeps the controller overhead independent of the number of windows.

The task register has no two-dimensional patch coordinate. Backbones with absolute patch positions leave the register without a patch positional embedding. In ViTDet global attention, decomposed relative-position terms apply only to patch-patch pairs, and entries involving the register are set to zero. This convention preserves the original patch geometry and introduces no positional parameters.

\subsection{Parameter and recovery-buffer cost}

For ViT-B with $d=768$, TAP-J adds one task register and two scalar readouts. Their total is
\begin{equation}
d+(d+1)+(d+1)=3d+2=2{,}306.
\end{equation}
The selection score introduces no new projection because it reuses the block query and key matrices. TAP-F applies rank-8 low-rank updates to the fused query-key-value projection, the attention output projection, and the two MLP linear maps in every transformer block. For ViT-B, one block adds
\begin{equation}
8\big[(d+3d)+(3d+d)+(d+4d)+(4d+d)\big]=98{,}304
\end{equation}
parameters. Across 12 blocks, the updates contain $1{,}179{,}648$ parameters. Adding the $768$-dimensional task register gives $1{,}180{,}416$ task-specific parameters, reported as $1.2$M. The shared allocation and recovery readouts are stored once and are not charged per task.

The TAP-J total of 2,306 counts the common dense-capable implementation, including a recovery readout. Classification does not call this readout. Removing it from a classification-only deployment reduces the add-on to $2d+1=1{,}537$ parameters, although the main tables retain the common 2.3K accounting convention.

The recovery buffer stores one $d$-dimensional offset and one pointer for each removed token, plus one scale per pruning layer. Its size is
\begin{equation}
M_{\mathrm{buffer}}=R(db_s+b_p)+Mb_s \quad \text{bytes},
\end{equation}
where $b_s$ is the scalar width and $b_p$ is the pointer width. Offsets and recovery scales use the active 16-bit activation type, which is BF16 in the reported profiles, and pointers use INT32. Records are indexed by original patch position, so recovery does not store a second spatial index for each removed token. The expression excludes temporary matching tensors and framework overhead. Section~\ref{sec:efficiency} compares the theoretical buffer with measured peak memory.
The main paper reports the dominant per-token term and omits the negligible $Mb_s$ layer-scale term.

\begin{table}[t]
\centering
\begin{tabular}{lrr}
\toprule
Component & TAP-J & TAP-F per task \\
\midrule
Initial task register & $d$ & $d$ \\
Allocation readout & $d+1$ & shared \\
Recovery readout & $d+1$ & shared \\
LoRA updates & 0 & 1,179,648 \\
Pruning-specific total & 2,306 & 1,180,416 \\
\bottomrule
\end{tabular}
\caption{Parameter accounting for ViT-B. Existing task heads are excluded from the pruning add-on, following the storage convention in the main paper.}
\label{tab:parameters}
\end{table}

\FloatBarrier
\section{Implementation and Reproducibility}
\label{sec:implementation}

\subsection{Shared protocol}

All task pipelines start from the same MAE-pretrained ViT-B family~\citep{mae}. We use ImageNet-1K for classification, ADE20K for semantic segmentation, and COCO for joint object detection and instance segmentation~\citep{imagenet,ade20k,coco}. Classification uses a linear head, segmentation uses a SegFormer MLP head, and detection uses ViTDet with Cascade Mask R-CNN~\citep{segformer,vitdet,cascade}. Detection and instance segmentation share one task register because one detector pipeline optimizes both outputs.

The three heads read the sparse backbone differently. Classification uses the final class token and does not invoke recovery. Segmentation reads blocks 3, 6, 9, and 12. TAP reconstructs the full grid independently at each read point before the four features enter the SegFormer head. Detection reconstructs the final backbone grid once, immediately before the simple feature pyramid and Cascade Mask R-CNN head. A reconstructed feature is never used as input to a later transformer block.

Table~\ref{tab:training-config} specifies the task-native pipelines used for the no-pruning checkpoints and TAP-J. Batch sizes are global across all workers. Throughput and memory use the separate single-GPU protocol in Section~\ref{sec:efficiency}.

\begin{table*}[t]
\centering
\begin{tabular}{llll}
\toprule
Setting & ImageNet-1K & ADE20K & COCO \\
\midrule
Input resolution & $224^2$ & $512^2$ & $1024^2$ \\
Patch grid & $14\times14$ & $32\times32$ & $64\times64$ \\
Training batch size & 1024 & 16 & 64 \\
Optimizer & AdamW & AdamW & AdamW \\
Base learning rate & $5\times10^{-4}$ & $6\times10^{-5}$ & $10^{-4}$ \\
Weight decay & 0.05 & 0.01 & 0.10 \\
Training length & 100 epochs & 160k iterations & 100 epochs \\
Warm-up and schedule & 5 epochs, cosine & 1500 iterations, polynomial & 250 iterations, steps at 88/96 \\
Augmentation & RRC, RA, Mixup, CutMix, RE & scale, crop, flip, photo & LSJ and horizontal flip \\
Pruning layers & 3, 6, 9 & 3, 6, 9 & 3, 6, 9 \\
Number of seeds & 3 & 5 & 3 \\
Primary metric & top-1 & mIoU & box AP \\
Secondary metric & n/a & n/a & mask AP \\
\bottomrule
\end{tabular}
\caption{Task-native configurations for the no-pruning checkpoints and TAP-J. RRC, RA, RE, and LSJ denote random resized crop, RandAugment, random erasing, and large-scale jittering. Detection uses one-based block indices. Its controlled probe also evaluates the final global-attention block.}
\label{tab:training-config}
\end{table*}

\subsection{TAP-J and TAP-F ownership}

The allocation and recovery readouts have no task-indexed branches and are reused across layers. Task dependence enters through the active register. TAP-J learns these components within each separately trained task checkpoint. TAP-F shares one allocation readout across all tasks and one recovery readout across the dense tasks.

Each forward pass contains exactly one active register, selected by the task identifier before the image enters the backbone. Switching tasks replaces that register and, in TAP-F, activates the corresponding low-rank updates and task head while the frozen backbone remains shared. The shared readouts receive the active register state and produce decisions that depend on both the task and the current image.

For each task, TAP-J trains a separate backbone, task register, allocation readout, task head, and, when needed, recovery readout. TAP-F freezes one shared backbone and trains the remaining modules for 480,000 optimizer steps. Each step samples one task uniformly, activates its register, head, and rank-8 LoRA updates, and performs one update with a shared AdamW optimizer~\citep{lora}. Inactive task modules receive no update on that step. This schedule gives each task about 160,000 updates, and the three data loaders cycle independently with their task-native augmentations. We train each TAP-F setting with five joint seeds. Segmentation summaries use all five runs, while classification and detection use the prespecified three-run subset that matches their task-native comparisons. We record the sampler seed, optimization seed, and realized task counts for every run. TAP-F uses a base learning rate of $10^{-4}$, weight decay 0.05, 5,000 warm-up steps, and cosine decay over the joint run. The low-rank updates target the fused query-key-value projection, attention output projection, and both MLP linear maps in every block. All tasks update the shared allocation readout, while only segmentation and detection update the recovery readout.

\begin{table}[H]
\centering
{\small
\begin{tabular}{@{}p{0.24\columnwidth}p{0.30\columnwidth}p{0.32\columnwidth}@{}}
\toprule
Component & TAP-J & TAP-F \\
\midrule
Base backbone & trainable per task & shared and frozen \\
Task register & trainable per task & trainable per task \\
LoRA updates & none & trainable per task \\
Allocation readout & trained per model & shared \\
Recovery readout & trained for dense task & shared by dense tasks \\
Task head & trainable per task & trainable per task \\
\bottomrule
\end{tabular}
}
\caption{Ownership and optimization of model components. ``Shared'' refers to one parameter set updated by the relevant task batches.}
\label{tab:ownership}
\end{table}

\subsection{Baseline provenance and common recovery}

Each baseline uses an archived implementation, checkpoint, and configuration. Dagger-marked dense baselines lack a native readout and use the common parameter-free reconstruction. This path matches each removed token to the surviving key with the highest cosine similarity. We use a held-out training subset to select one fixed endpoint per task. Segmentation uses the full offset, and detection uses the stand-in. This subset is disjoint from benchmark validation.

Table~\ref{tab:ownership} distinguishes parameter sharing from experimental comparability. TAP-F shares the frozen base and controller readouts, but each task owns the modules that define its interface to that base. Tables~\ref{tab:baseline-provenance} and~\ref{tab:dense-baseline-provenance} record implementation provenance, native task pipelines, and the use of common dense recovery.

The provenance record supports comparisons at two levels. Within a task, every method uses the same dataset, input resolution, metric, and throughput protocol, giving a deployment-level comparison of complete pipelines. Across tasks, the study asks whether one pruning design remains useful as the objective and prediction interface change. Methods retain their native readouts. Common recovery supplies missing dense positions only for baselines that define no readout.

\begin{table}[H]
\centering
{\footnotesize
\begin{tabular}{@{}p{0.21\columnwidth}p{0.28\columnwidth}p{0.36\columnwidth}@{}}
\toprule
Method & Source & Classification path \\
\midrule
EViT & official~\citep{evit} & token fusion, official setup \\
DynamicViT & official~\citep{dynamicvit} & learned predictors, official setup \\
Evo-ViT & official~\citep{evovit} & slow-fast update, official setup \\
ATS & official~\citep{ats} & attention sampling, official setup \\
ToMe & official~\citep{tome} & task-native configuration \\
DiffRate & official~\citep{diffrate} & task-native configuration \\
Token Cropr & official~\citep{tokencropr} & task-specific official setup \\
\bottomrule
\end{tabular}
}
\caption{Classification baseline provenance. The anonymous code manifest records the archived repository commit, configuration, checkpoint checksum, and local changes for every row.}
\label{tab:baseline-provenance}
\end{table}

\begin{table}[!t]
\centering
{\footnotesize
\begin{tabular}{@{}p{0.20\columnwidth}p{0.25\columnwidth}p{0.39\columnwidth}@{}}
\toprule
Method & Source & Dense readout and adaptation \\
\midrule
Attn Top-K & EViT score~\citep{evit} & no native readout, common recovery \\
ToMe & official~\citep{tome} & task-dependent readout, task-native setup \\
DiffRate & official~\citep{diffrate} & no native readout, common recovery on ADE20K \\
ALGM & official~\citep{algm} & native readout, official dense pipeline \\
DToP & official~\citep{dtop} & native readout, official dense pipeline \\
Token Cropr & official~\citep{tokencropr} & native readout, task-specific official setup \\
SViT & official~\citep{svit} & native readout, official detector setup \\
\bottomrule
\end{tabular}
}
\caption{Dense-task baseline provenance and readout handling. Common recovery uses the task endpoint selected on the training calibration subset.}
\label{tab:dense-baseline-provenance}
\end{table}

The common path is parameter-free. It reuses each method's retained set, matches removed positions with the same key-space rule, and fixes the task endpoint on the calibration subset. Methods with native recovery keep their published mechanism. This separation lets the main tables report pruning quality, dense readout design, and task-head cost as distinct parts of each pipeline.

\section{Controlled-Probe Study}
\label{sec:probes}

\subsection{Protocol}

Each probe freezes the backbone and head of the corresponding no-pruning checkpoint. It changes one eligible layer, retains the same hard token count, and evaluates the resulting pruned model without retraining. The criteria are parameter-free functions of the frozen states. They span incoming attention, activation magnitude, spatial coverage, and pairwise feature affinity. Learned selectors are excluded from the controlled probe and appear only as trained baselines in the main comparisons.

\begin{figure*}[!t]
\centering
\includegraphics[width=\textwidth]{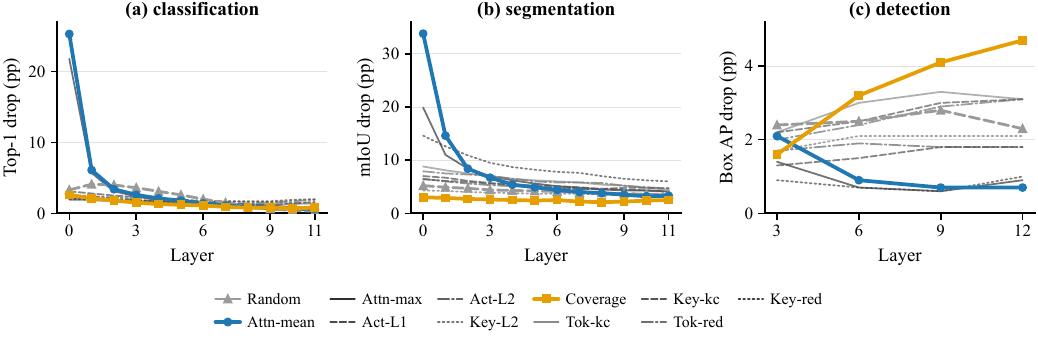}
\caption{Complete layerwise controlled probes. Each curve changes one criterion at one eligible layer of the frozen task checkpoint. Attention, coverage, and random selection use the colors and markers of main-paper Figure 1. The remaining criteria are shown in grey. Lower values are better. Detection is evaluated at global-attention blocks 3, 6, 9, and 12.}
\label{fig:all-probes}
\end{figure*}

The criterion study uses one designated no-pruning checkpoint for each task because the probe itself has no training stage. The paired recovery study repeats the evaluation over the checkpoint seeds listed in Table~\ref{tab:training-config}. Its standard deviation is computed from within-seed differences between the two recovery endpoints.

The diagnostic attention criterion is the incoming patch-attention probability averaged over heads and query tokens. It is measured from the frozen full-attention block and is not used as an efficient pre-attention selector. This definition also applies to pipelines without a class token. Coverage runs deterministic farthest point sampling after scaling both patch-centre coordinates to $[0,1]$ and begins at the patch nearest the image centre~\citep{fps}. Affinity criteria retain one representative and discard its paired token, preserving the same hard count as pruning. Each task keeps its native resolution, head, and eligible layers. The resulting differences compare complete pipelines and do not isolate task semantics from pipeline design.

\subsection{Criterion definitions and layerwise values}

Table~\ref{tab:criteria} gives the complete reference set. The main paper groups the criteria and highlights attention, coverage, and random selection so Figure 1 remains readable. All data-dependent scores are computed from the same frozen pre-attention states. Token and key k-centre selection starts from the candidate closest to the corresponding feature mean. Every deterministic tie is resolved by original patch index.

The reference set also clarifies what it means to prune a shared ViT backbone. Parameter sharing does not require every task to assign the same value to a patch. The frozen probes expose all criteria to states produced by the same MAE-pretrained model family, but their relative quality changes with the head, resolution, eligible layers, and objective. One common representation can support several task-dependent orderings. The backbone supplies the token space in which relevance is measured, while the active task determines which evidence must survive.

Depth adds a second source of variation. Early tokens retain local position and texture that later blocks may combine into broader semantic evidence. Removing a spatially concentrated set at this stage can erase information that no later block can recreate. Near the output, a task may benefit from concentrating computation on regions already identified as useful. Figure~\ref{fig:all-probes} shows this transition directly. Attention is weakest in the earliest classification and segmentation layers, but the relation changes deeper in the network and differs again for detection. A unified backbone can share its feature transformations, but the route through them still needs a task-aware and depth-aware control signal.

The criteria in Table~\ref{tab:criteria} also test different notions of evidence. Attention and activation magnitude ask which individual patches appear strong in the current state. Coverage asks whether the retained set still spans the image. The affinity criteria instead search for patches whose information may already be represented elsewhere. These questions are not equivalent in a shared feature space. A high-magnitude patch may be redundant, and a weak local response may still mark the only surviving evidence for a boundary or a small object. Their relative value can change after each block because attention mixes evidence before the next pruning site. The controlled study evaluates complete retained sets at fixed cardinality rather than interpreting a score in isolation.

Resolution and readout geometry further change what that fixed cardinality represents. A classification head consumes one global summary, while a dense head must recover evidence across many spatial positions. Increasing the patch grid also creates more locally similar candidates, which can make redundancy useful without making spatial coverage irrelevant. Keeping each pipeline's native grid and head lets the probes test whether a criterion preserves its ordering under the full conditions in which pruning operates.

Pruning also changes the states that drive later decisions. An early removal eliminates the current patch feature and its future interactions, altering the evidence available at the next eligible layer. This matters especially for segmentation because its head reads several backbone depths. An early selection error affects the feature tapped at that depth and the context passed to later taps. Detection reads the final grid after further task-specific processing, which allows object-centred evidence to concentrate before recovery. The single-layer probes measure this local sensitivity, and their depth trends specify the information that the learned allocator must combine when TAP prunes at several layers.

\begin{table}[H]
\centering
{\footnotesize
\begin{tabular}{@{}p{0.22\columnwidth}p{0.17\columnwidth}p{0.48\columnwidth}@{}}
\toprule
Criterion & Family & Rule and input \\
\midrule
Random & reference & stateless seeded permutation keyed by sample and layer \\
Attention mean & attention & retain the largest mean incoming attention from the frozen block matrix \\
Attention max & attention & retain the largest maximum incoming attention from the frozen block matrix \\
Activation $\ell_1$ & magnitude & retain the largest pre-LN token $\ell_1$ norm \\
Activation $\ell_2$ & magnitude & retain the largest pre-LN token $\ell_2$ norm \\
Key $\ell_2$ & magnitude & retain the largest projected-key norm \\
Coverage & spatial & farthest point sampling from the image centre using patch centres \\
Token k-centre & diversity & farthest point sampling by normalized-token cosine distance \\
Key k-centre & diversity & farthest point sampling by normalized-key cosine distance \\
Token redundancy & affinity & greedily remove the more redundant member of the closest normalized-token pair \\
Key redundancy & affinity & greedily remove the more redundant member of the closest normalized-key pair \\
\bottomrule
\end{tabular}
}
\caption{Reduction criteria in the complete controlled study. Dense tasks use the paired recovery setting. For a redundancy criterion, the more redundant member has the larger mean cosine similarity to the current set. An exact tie removes the larger original patch index.}
\label{tab:criteria}
\end{table}

Table~\ref{tab:criterion-summary} tests whether a criterion transfers when its layerwise drops are averaged within each pipeline. Classification uses top-1 accuracy, segmentation uses mIoU, and detection uses box AP.

\begin{table}[H]
\centering
{\footnotesize
\begin{tabular}{@{}p{0.32\columnwidth}rrr@{}}
\toprule
Criterion & Cls & Seg (rank) & Det (rank) \\
\midrule
Random & 2.2 & 4.2 (3) & 2.5 (7) \\
Attention mean & 3.9 & 8.0 (10) & 1.1 (3) \\
Attention max & 3.4 & 7.1 (9) & 0.9 (2) \\
Activation $\ell_1$ & 1.8 & 5.3 (6) & 1.6 (4) \\
Activation $\ell_2$ & 1.6 & 5.0 (4) & 1.8 (5) \\
Key $\ell_2$ & 1.9 & 3.8 (2) & 2.0 (6) \\
Coverage & 1.3 & 2.5 (1) & 3.4 (11) \\
Token k-centre & 1.5 & 6.3 (8) & 2.9 (10) \\
Key k-centre & 1.6 & 5.2 (5) & 2.7 (9) \\
Token redundancy & 1.7 & 6.2 (7) & 2.6 (8) \\
Key redundancy & 1.8 & 8.8 (11) & 0.8 (1) \\
\bottomrule
\end{tabular}
}
\caption{Mean single-layer drops in percentage points across eligible layers. Lower values and ranks are better. Across the 11 criteria, the Spearman correlation between the two dense-task rankings is $-0.62$.}
\label{tab:criterion-summary}
\end{table}

Coverage ranks first for segmentation but last for detection. Key redundancy shows the opposite pattern, and the full dense-task rankings have a Spearman correlation of $-0.62$. Figure~\ref{fig:all-probes} shows where these averages arise. Early classification and segmentation layers account for much of the attention loss, while detection reverses the relative ordering of attention and coverage after its first eligible block.

\FloatBarrier
\raggedbottom

\subsection{Paired recovery probe}

The recovery probe uses coverage at a local keep rate of $0.5$. Each pair shares the retained set, stand-in pointers, stored offsets, and checkpoint. The only change is the endpoint: full-offset recovery sets $\alpha=1$, and stand-in-only recovery sets $\alpha=0$. Pairing removes selection variation from the comparison. Table~\ref{tab:paired-recovery} reports every eligible layer and the paired difference.

\begin{table}[H]
\centering
{\footnotesize
\begin{tabular}{@{}llrrr@{}}
\toprule
Task & Layer & Full offset & Stand-in & Paired $\Delta\pm$SD \\
\midrule
Seg & 3 & 46.2 & 44.6 & $+1.6\pm0.12$ \\
Seg & 6 & 46.0 & 44.2 & $+1.8\pm0.15$ \\
Seg & 9 & 45.5 & 43.5 & $+2.0\pm0.18$ \\
Det box & 3 & 51.8 & 52.9 & $-1.1\pm0.08$ \\
Det box & 6 & 51.6 & 52.8 & $-1.2\pm0.09$ \\
Det box & 9 & 51.3 & 52.5 & $-1.2\pm0.10$ \\
Det box & 12 & 50.9 & 52.2 & $-1.3\pm0.12$ \\
Det mask & 3 & 44.8 & 45.8 & $-1.0\pm0.07$ \\
Det mask & 6 & 44.6 & 45.6 & $-1.0\pm0.08$ \\
Det mask & 9 & 44.3 & 45.4 & $-1.1\pm0.09$ \\
Det mask & 12 & 43.9 & 45.1 & $-1.2\pm0.11$ \\
\bottomrule
\end{tabular}
}
\caption{Paired recovery probe. The no-pruning anchors are 47.2 mIoU, 54.0 box AP, and 46.7 mask AP. Segmentation uses five seeds and detection uses three. A positive difference favors the full offset. SD is computed across paired seed differences with identical retained sets, pointers, and checkpoints.}
\label{tab:paired-recovery}
\end{table}

\section{Extended Results}
\label{sec:extended-results}

\subsection{Verified operating point at $\rho=0.5$}

Table~\ref{tab:verified-main} collects the TAP operating points stated in the main paper. The no-pruning deltas and throughput multipliers agree with its Tables 1--3 and anchor the additional keep-rate results below.

\begin{table}[H]
\centering
{\footnotesize
\begin{tabular}{@{}llrrrr@{}}
\toprule
Task & Regime & Score & $\Delta$ & Enc. im/s & Speedup \\
\midrule
Cls & TAP-J & 83.2 & $-0.4$ & 2300 & $1.26\times$ \\
Cls & TAP-F & 82.6 & $-1.0$ & 2288 & $1.25\times$ \\
Seg & TAP-J & 47.0 & $-0.2$ & 280 & $1.30\times$ \\
Seg & TAP-F & 45.8 & $-1.4$ & 275 & $1.27\times$ \\
Det & TAP-J & 53.7 & $-0.3$ & 18.8 & $1.32\times$ \\
Det & TAP-F & 52.5 & $-1.5$ & 18.4 & $1.30\times$ \\
\bottomrule
\end{tabular}
}
\caption{Verified TAP results at $\rho=0.5$. Cls, Seg, and Det denote ImageNet-1K classification, ADE20K segmentation, and COCO box AP. COCO full-model throughput is 11.5 im/s for TAP-J and 11.3 im/s for TAP-F.}
\label{tab:verified-main}
\end{table}

\subsection{Keep-rate results}

Table~\ref{tab:keep-rates} reports $\rho\in\{0.3,0.5,0.7\}$ for TAP-J and TAP-F. We train a separate model at each keep rate and hold all other settings fixed within each task and adaptation regime. Scores are means across five seeds for segmentation and three seeds for the other tasks. The table gives the sample standard deviation, encoder throughput, full-model throughput where applicable, and peak memory. Full-model throughput is listed only for detection to match the main comparison, while Table~\ref{tab:latency-breakdown} reports end-to-end latency for all tasks at $\rho=0.5$.

At $\rho=0.5$, TAP-J remains within 0.1 point of its $\rho=0.7$ score on all three tasks and raises encoder throughput by $7$--$11\%$. TAP-F makes a larger score trade-off, losing 0.3 top-1 points, 0.6 mIoU, and 0.5 box AP for similar throughput gains. These results support $\rho=0.5$ as the balanced operating point used in the main paper.

\begin{table}[H]
\centering
{\footnotesize
\setlength{\tabcolsep}{2.5pt}
\begin{tabular}{@{}lrrrrr@{}}
\toprule
Setting & $\rho$ & Score $\pm$ SD & Enc. & Full & GB \\
\midrule
Cls J & 0.3 & $83.1\pm0.09$ & 2683 & n/a & 0.35 \\
Cls J & 0.5 & $83.2\pm0.06$ & 2300 & n/a & 0.37 \\
Cls J & 0.7 & $83.5\pm0.04$ & 2154 & n/a & 0.39 \\
Cls F & 0.3 & $82.1\pm0.12$ & 2669 & n/a & 0.36 \\
Cls F & 0.5 & $82.6\pm0.08$ & 2288 & n/a & 0.38 \\
Cls F & 0.7 & $82.9\pm0.06$ & 2141 & n/a & 0.40 \\
Seg J & 0.3 & $46.3\pm0.29$ & 330 & n/a & 0.58 \\
Seg J & 0.5 & $47.0\pm0.18$ & 280 & n/a & 0.64 \\
Seg J & 0.7 & $47.1\pm0.13$ & 254 & n/a & 0.70 \\
Seg F & 0.3 & $44.7\pm0.34$ & 325 & n/a & 0.61 \\
Seg F & 0.5 & $45.8\pm0.24$ & 275 & n/a & 0.66 \\
Seg F & 0.7 & $46.4\pm0.20$ & 250 & n/a & 0.72 \\
Det J & 0.3 & $52.8\pm0.19$ & 22.2 & 12.3 & 2.70 \\
Det J & 0.5 & $53.7\pm0.11$ & 18.8 & 11.5 & 2.82 \\
Det J & 0.7 & $53.8\pm0.08$ & 16.9 & 10.6 & 2.94 \\
Det F & 0.3 & $51.2\pm0.23$ & 21.8 & 12.1 & 2.77 \\
Det F & 0.5 & $52.5\pm0.14$ & 18.4 & 11.3 & 2.88 \\
Det F & 0.7 & $53.0\pm0.16$ & 16.5 & 10.4 & 3.00 \\
\bottomrule
\end{tabular}
}
\caption{Keep-rate results for TAP-J (J) and TAP-F (F). Score denotes top-1 accuracy, mIoU, or box AP. Enc. and Full are images per second. GB is peak memory.}
\label{tab:keep-rates}
\end{table}

\section{Ablations}
\label{sec:ablations}

\subsection{Absolute values at $\rho=0.5$}

The main paper reports changes from TAP-F. Table~\ref{tab:absolute-ablation} converts those changes to absolute metrics using the same rounded reference values so the rows can be compared directly.

\begin{table}[H]
\centering
\begin{tabular}{lrrr}
\toprule
Variant & Cls & Seg & Det \\
\midrule
TAP-F full & 82.6 & 45.8 & 52.5 \\
Attention selection & 82.4 & 44.9 & 52.0 \\
Static task query & 82.5 & 45.4 & 52.2 \\
Recovery off & n/a & 42.4 & 50.3 \\
$\alpha=1$ & n/a & 45.7 & 51.5 \\
$\alpha=0$ & n/a & 44.2 & 52.4 \\
Task adapters, shared register & 82.4 & 44.5 & 51.4 \\
Task-shared allocation & 82.5 & 45.2 & 52.0 \\
Layer-uniform allocation & 82.0 & 44.6 & 51.7 \\
Task-mean allocation & 82.5 & 45.6 & 52.4 \\
Task-mean recovery & n/a & 45.7 & 52.4 \\
Both task means & 82.5 & 45.5 & 52.3 \\
\bottomrule
\end{tabular}
\caption{Absolute metrics for the ablations in main-paper Table 4, using the same rounded TAP-F reference values.}
\label{tab:absolute-ablation}
\end{table}

\subsection{Controls and reporting protocol}

Each ablation changes one component while fixing the backbone, adapters, token count, training budget, and seeds. Register controls separate task identity from feature adaptation because a shared initial register may still receive task-specific adapter features. Allocation controls record every layer count and verify the exact-budget identity for each image. Recovery endpoint controls reuse the same retained sets and pointers. Replacing a per-image output with its task-layer mean leaves all other components unchanged.

\subsection{Separating task identity from feature adaptation}

The main ablation shares the initial register but retains task-specific adapters, changing only one axis of the design. A complete test crosses register ownership with adapter ownership. The four models use the same frozen base, heads, optimizer, task sampler, keep rate, seeds, and training length. Task identity selects the task-specific register before the forward pass. A shared register uses one learned initial vector for all task batches. Task-specific adapters activate a separate low-rank update for each task, while the shared-adapter condition trains one low-rank update on the mixed task stream. Table~\ref{tab:factorial} isolates whether the register remains useful when the features are not already separated by task-specific adapters.

\begin{table}[H]
\centering
\begin{tabular}{llrrr}
\toprule
Register & Adapter & Cls & Seg & Det \\
\midrule
task-specific & task-specific & 82.6 & 45.8 & 52.5 \\
shared & task-specific & 82.4 & 44.5 & 51.4 \\
task-specific & shared & 81.9 & 44.7 & 51.8 \\
shared & shared & 81.2 & 42.8 & 49.9 \\
\bottomrule
\end{tabular}
\caption{Register-by-adapter factorial at $\rho=0.5$. All entries are absolute metrics. The first two rows correspond to the full model and shared-register ablation in the main paper.}
\label{tab:factorial}
\end{table}

Sharing the adapter lowers all three metrics even when the register remains task-specific. Sharing both components increases the losses to 1.4 top-1 points, 3.0 mIoU, and 2.6 box AP relative to TAP-F. The interaction is strongest on the dense tasks, where the register and low-rank path provide complementary forms of task specialization.

\subsection{Stand-in matching controls}

Recovery depends on two coupled choices. Matching selects the surviving feature and defines the stored offset. The endpoint scale determines how much of that offset is restored. Endpoint comparisons alone do not test the matching rule. We keep the selected token set, allocation, recovery readout, checkpoint, and random seeds fixed, then replace only the stand-in assignment. The controls use cosine similarity in token space, Euclidean distance on the patch grid, the highest-scoring retained token, and a seeded random retained token. Pointer chains and original-index tie handling remain unchanged. A separate control resolves every removed position directly against the final survivor set. It tests whether the chain itself, rather than key-space similarity, accounts for the result.

Table~\ref{tab:matching-ablation} reports the effect of changing only the stand-in assignment. Table~\ref{tab:matching-recovery} then crosses each matching rule with the three recovery endpoints without changing the selected token sets or layer allocations. This paired sweep tests whether a stronger match changes the amount of offset preferred by each task.

\begin{table}[H]
\centering
{\small
\begin{tabular}{@{}lrrrr@{}}
\toprule
Matching rule & Seg & $\Delta$ & Det & $\Delta$ \\
\midrule
Key cosine, pointer chain & 45.8 & 0.0 & 52.5 & 0.0 \\
Token cosine & 45.6 & $-0.2$ & 52.4 & $-0.1$ \\
Spatial nearest & 45.1 & $-0.7$ & 51.9 & $-0.6$ \\
Highest-score stand-in & 44.7 & $-1.1$ & 51.6 & $-0.9$ \\
Random stand-in & 43.5 & $-2.3$ & 50.7 & $-1.8$ \\
Direct final-survivor match & 45.4 & $-0.4$ & 52.2 & $-0.3$ \\
\bottomrule
\end{tabular}
}
\caption{Stand-in controls in TAP-F at $\rho=0.5$. Changes are measured from key-cosine matching with pointer chains.}
\label{tab:matching-ablation}
\end{table}

\begin{table}[H]
\centering
{\footnotesize
\setlength{\tabcolsep}{3pt}
\begin{tabular}{@{}lrr@{}}
\toprule
Matching rule & Seg $0/L/1$ & Det $0/L/1$ \\
\midrule
Key cosine, pointer chain & 44.2/45.8/45.7 & 52.4/52.5/51.5 \\
Token cosine & 43.9/45.6/45.3 & 52.2/52.4/51.1 \\
Spatial nearest & 43.1/45.1/44.4 & 51.8/51.9/50.6 \\
Highest-score stand-in & 42.7/44.7/43.8 & 51.5/51.6/50.2 \\
Random stand-in & 41.6/43.5/42.4 & 50.4/50.7/48.9 \\
Direct final-survivor & 43.7/45.4/45.1 & 52.0/52.2/50.8 \\
\bottomrule
\end{tabular}
}
\caption{Interaction between stand-in matching and recovery in TAP-F at $\rho=0.5$. Within each task, entries give $\alpha=0$, the learned scale (L), and $\alpha=1$. Selected sets, allocations, checkpoints, and seeds are fixed.}
\label{tab:matching-recovery}
\end{table}

Key-space matching gives the best result on both dense tasks, and token-space cosine remains close. Spatial, score-based, and random stand-ins lose more, showing that a useful endpoint requires more than proximity or selector rank. Direct matching to final survivors also trails the pointer chain. Across all matching rules, segmentation favors the learned or full-offset endpoint. Detection favors the learned or stand-in-heavy endpoint.

\subsection{Training estimator controls}

The constrained-ST variants and the hard-only control use the same initialization and sparse forward path. They differ only in the gradient estimator or its temperature. In the hard-only control, discrete decisions block the task-loss gradient to the scoring and allocation readouts. The independent-sigmoid baseline instead thresholds separate gates and controls their mean with a rate penalty, so its per-image sequence length may vary. Every row reports its realized final token count in addition to task performance. This comparison separates the exact inference budget from the mechanism used to train selection and allocation.

\begin{table}[!b]
\centering
{\footnotesize
\setlength{\tabcolsep}{3pt}
\begin{tabular}{@{}p{0.32\columnwidth}rrr@{}}
\toprule
Estimator & Cls/Seg/Det & Count & Exact \\
\midrule
Constrained ST, cosine $\tau$ & 82.6/45.8/52.5 & $K\pm0$ & yes \\
No Gumbel perturbation & 82.5/45.4/52.2 & $K\pm0$ & yes \\
Fixed $\tau=1.0$ & 82.2/44.9/51.9 & $K\pm0$ & yes \\
Independent sigmoid and rate loss & 81.7/44.3/51.2 & $1.014K\pm0.053K$ & no \\
Hard selection, no surrogate & 80.8/41.9/49.4 & $K\pm0$ & yes \\
\bottomrule
\end{tabular}
}
\caption{Training-estimator controls at $\rho=0.5$. Count reports the mean and standard deviation of the final token count. ``Exact'' states whether every image reaches the prescribed integer count without a rate tolerance.}
\label{tab:training-ablation}
\end{table}

Removing Gumbel perturbations has a small cost, while a fixed high temperature produces a larger drop on every task. Independent sigmoid gates miss the prescribed count and reduce all three metrics. Hard selection without a surrogate preserves the count but causes the largest loss, confirming that the controller needs a gradient through its discrete decisions.

\section{What the Task Register Learns}
\label{sec:register-analysis}

\begin{figure}[H]
\centering
\includegraphics[width=0.98\columnwidth]{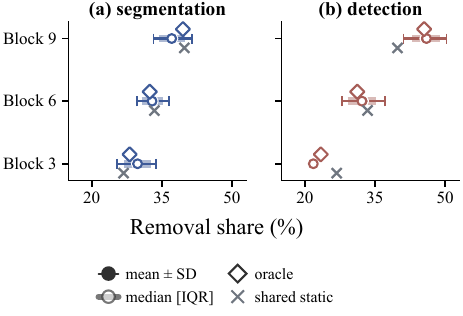}
\caption{Per-image allocation summaries for TAP-F at $\rho=0.5$. Thin intervals show mean $\pm$ SD, thick intervals show the median and IQR, diamonds mark the calibration-grid oracle, and crosses mark the shared static schedule from main-paper Figure 5(a). Every image retains the same final token count.}
\label{fig:allocation-distributions}
\end{figure}

\begin{figure}[H]
\centering
\includegraphics[width=0.98\columnwidth]{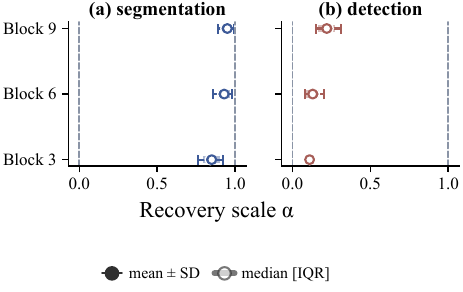}
\caption{Per-image recovery-scale summaries for TAP-F at $\rho=0.5$. Thin intervals show mean $\pm$ SD and thick intervals show the median and IQR. The dashed lines mark stand-in-only recovery at $\alpha=0$ and full-offset recovery at $\alpha=1$.}
\label{fig:recovery-distributions}
\end{figure}

The main paper reports validation-set means for allocation and recovery. Means alone do not show whether every image receives a similar policy or whether the task mean hides substantial image variation. The summaries here use the designated TAP-F checkpoint at $\rho=0.5$ and all images in the corresponding validation split. Table~\ref{tab:register-distributions} reports the standard deviation, median, and interquartile range across images for each dense task and pruning layer. Figure~\ref{fig:allocation-distributions} and Figure~\ref{fig:recovery-distributions} show the corresponding distributions.

Segmentation updates its register before the first pruning layer, so all three decisions may vary by image. Detection follows a different schedule. Its register bypasses the two window-attention blocks that precede block 3, which leaves the first allocation and recovery scale task-conditioned but constant across images. After block 3 provides the first global interaction, the register can adapt the decisions at blocks 6 and 9 to the current image.

The diagnostic calibration-grid oracle evaluates a prespecified set of legal layer splits for every image in held-out subsets drawn from the training data. Each candidate removal share is a multiple of $5\%$, which gives 231 nonnegative triplets that sum to $100\%$. Largest-remainder conversion maps every triplet to integer counts with the same exact total budget, using earlier layers to resolve equal remainders. The calibration subsets contain 500 ADE20K images and 1,000 COCO images and remain disjoint from the reported validation sets. For each image, the native task loss selects the candidate: pixelwise cross-entropy for segmentation and the detector's classification-plus-box loss for detection. We average the winning splits and evaluate that fixed mean schedule on the validation set. Table~\ref{tab:allocation-controls} compares it with the learned policy and simpler static schedules while holding selection, recovery, checkpoints, and seeds fixed.

\begin{table}[H]
\centering
{\footnotesize
\setlength{\tabcolsep}{2.2pt}
\begin{tabular}{@{}lrrrr@{}}
\toprule
Policy & Seg split & Seg & Det split & Det \\
\midrule
Learned per image & 29.6/33.1/37.3 & 45.8 & 21.8/32.5/45.7 & 52.5 \\
Task-mean static & 29.6/33.1/37.3 & 45.6 & 21.8/32.5/45.7 & 52.4 \\
Calibration-grid oracle & 28.1/32.4/39.5 & 45.9 & 23.4/31.2/45.5 & 52.6 \\
Shared static & 26.8/33.4/39.8 & 45.2 & 26.8/33.4/39.8 & 52.0 \\
Layer uniform & 33/33/34 & 44.6 & 33/33/34 & 51.7 \\
Early heavy & 45/35/20 & 44.9 & 45/35/20 & 51.9 \\
Late heavy & 20/30/50 & 45.4 & 20/30/50 & 52.5 \\
\bottomrule
\end{tabular}
}
\caption{Allocation controls in TAP-F at $\rho=0.5$. Splits give the percentage of the removal budget assigned to blocks 3, 6, and 9. The learned row varies by image and reports its validation-set mean. The oracle row averages per-image winners on calibration data and evaluates the resulting fixed schedule on validation. The remaining rows also use one fixed schedule per task. Shared static reproduces main-paper Figure 5(a). Displayed split means may total $100.1\%$ after independent rounding.}
\label{tab:allocation-controls}
\end{table}

Table~\ref{tab:allocation-controls} separates image-dependent allocation from the mean schedule. Freezing the task mean costs 0.2 mIoU and 0.1 box AP, while a shared static schedule loses 0.6 and 0.5 points. The calibration oracle is only 0.1 point above the learned policy on either task. The comparison indicates that most of the benefit comes from learning a task-specific schedule, with a smaller gain from adapting that schedule to each image.

Allocation and recovery expose different parts of the register state. The allocation readout determines when computation is removed, so its outputs are coupled by the exact global budget. The recovery readout acts only after a removed position has been assigned a surviving endpoint. Its scale can change with task and depth without altering the sparse sequence. Figure~\ref{fig:recovery-distributions} tests whether this second readout collapses to one fixed endpoint or preserves image-dependent variation after training.

\begin{table}[!t]
\centering
{\footnotesize
\setlength{\tabcolsep}{2.5pt}
\begin{tabular}{@{}lrrrr@{}}
\toprule
& \multicolumn{2}{c}{Removal share (\%)} & \multicolumn{2}{c}{Recovery scale} \\
\cmidrule(lr){2-3}\cmidrule(lr){4-5}
Task/layer & Mean $\pm$ SD & Med. [IQR] & Mean $\pm$ SD & Med. [IQR] \\
\midrule
Seg 3 & $29.6\pm4.2$ & 29.8 [5.7] & $0.84\pm0.08$ & 0.85 [0.11] \\
Seg 6 & $33.1\pm3.5$ & 32.9 [4.5] & $0.92\pm0.06$ & 0.93 [0.08] \\
Seg 9 & $37.3\pm4.1$ & 37.1 [5.6] & $0.94\pm0.05$ & 0.95 [0.06] \\
Det 3 & $21.8\pm0.0$ & 21.8 [0.0] & $0.11\pm0.00$ & 0.11 [0.00] \\
Det 6 & $32.5\pm4.6$ & 32.2 [6.1] & $0.14\pm0.06$ & 0.13 [0.08] \\
Det 9 & $45.7\pm4.6$ & 46.0 [6.0] & $0.23\pm0.08$ & 0.22 [0.11] \\
\bottomrule
\end{tabular}
}
\caption{Allocation and recovery summaries at $\rho=0.5$. The means reproduce main-paper Figure 5. Removal shares sum to $100\%$ for every image. Detection block 3 is constant across images because the register has not yet participated in global attention.}
\label{tab:register-distributions}
\end{table}

Table~\ref{tab:register-distributions} tests whether the mean curves in the main paper hide image-dependent policies. Segmentation varies at all three layers. Detection is fixed at block 3 and redistributes the remaining budget between blocks 6 and 9 after the register has received global image context. The recovery statistics show a second form of specialization. Segmentation remains close to full-offset recovery, while detection stays near the stand-in endpoint and becomes more image-dependent only after the register has participated in global attention.

\flushbottom
\section{Sharing a ViT Backbone Across Tasks}
\label{sec:unified-backbone}

A unified visual backbone is often described only in terms of parameter reuse. That definition is incomplete for conditional computation. Two tasks may use the same transformer weights and still require different tokens, pruning depths, and dense outputs. TAP shares the representational machinery and sparse execution contract while allowing each task to follow a different path through the available computation.

\subsection{A common representation, not a common ranking}

The task register operates in the same $d$-dimensional residual space as the patch tokens. The block's existing query projection produces its selection query, and the corresponding key projection processes candidate patches. The policy reads the backbone in its native coordinates instead of introducing a selector with a separate feature space. A distinct predictor for every task could learn an effective ranking, but it would also create a new interface that must remain aligned with the backbone. TAP keeps this interface inside the transformer, and switching the active register changes how the same patch representation is queried.

This operational role distinguishes a task register from a task-agnostic storage token. A storage register can absorb recurring artifacts and improve the quality of the shared representation~\citep{registers}. The TAP register also controls computation. Its state is read before selected blocks and converted into decisions that change the amount and location of subsequent computation. The register remains compact because it does not carry a second feature pyramid or a bank of layer-specific selectors. Image evidence reaches it through ordinary transformer interactions, so the control state evolves with the same computation that it regulates.

TAP-F makes this parameter sharing explicit. It reuses one frozen base and gives each task a register and low-rank updates~\citep{lora}. The low-rank path can adapt the patch states, and each task retains its own head. Task adaptation becomes a bounded modification of one base instead of a separate dense backbone.

\subsection{Depth separates sharing from specialization}

A ViT block does not have a fixed semantic role independent of its position. Early blocks build local relations from patches whose spatial support is still explicit. Later blocks act on features that have accumulated context through several attention and feed-forward updates. Removing a token at block 3 forfeits different future interactions from removing one at block 9. The layerwise probes make this distinction visible, and the allocation distributions show how the register responds to it.

The exact global budget separates how much computation the backbone may use from where it is spent. The keep rate fixes the amount, and the register allocates it across depth. Because every image reaches the same final integer count, allocation differences reflect where the budget is spent. Segmentation tends to protect more early evidence, consistent with its need to preserve boundaries and semantic coverage. Detection spends more of the removal budget later, after global interactions make object-centred evidence easier to identify. These task-specific patterns show why depth remains part of the interface even when the transformer parameters are shared.

The detection schedule also reveals a limit of register conditioning. Before the first global-attention block, the register has not received global image context, so the first allocation and recovery scale are task-conditioned but constant across images. Token scores remain image-dependent because they use the current patch keys. Later decisions can depend on both the task and the image. A unified backbone must expose enough communication points for the controller to observe the information on which its decisions depend. Adding a register alone cannot make a locally isolated block globally informed.

\subsection{A sparse backbone with task-specific boundaries}

Dense prediction constrains how a shared sparse backbone can expose its output. A segmentation or detection head expects features at spatial positions that the sparse encoder no longer carries. Reinserting reconstructed tokens into the backbone would make later computation dense again and would mix task-specific recovery with the shared transformation. TAP instead treats recovery as a boundary operation. The encoder remains sparse, and a complete grid is formed only when the task head requests one.

This boundary is shared in form but not in value. Every removed token is matched to a surviving endpoint by the same key-space rule. The task register controls how much of the stored offset is restored at each eligible depth. Segmentation remains close to the full-offset endpoint because fine spatial distinctions remain useful at readout. Detection stays closer to the final stand-in because its head benefits more from the surviving object-centred representation. The same sparse backbone can support both behaviours because recovery does not impose either endpoint as a universal reconstruction rule.

These components provide a practical form of backbone sharing. A common ViT provides the token space, transformer stack, and exact compute budget. Task registers specialize the path through that stack, and dense recovery specializes the interface to the output head. A new task would require eligible pruning layers and a learned task interface, while the budget mechanism and sparse encoder contract remain unchanged. The present experiments cover three established pipelines. Scaling to a larger task set will require measuring interference among task interfaces and determining when low-rank adaptation remains sufficient.

\section{Efficiency, Throughput, and Memory}
\label{sec:efficiency}

\subsection{Profiling protocol}

The measurements use one 24GB NVIDIA RTX PRO 4000 Blackwell with BF16 autocast. Throughput batches contain 128 classification images, 8 segmentation images, or 1 detection image. Each run performs 100 warm-up iterations followed by 500 measured iterations. We synchronize CUDA before and after every measured region and exclude data loading and host-to-device transfer. All methods within a task use the same resolution, batch size, and procedure. TAP-F keeps its low-rank updates unmerged during profiling, matching the main tables.

\subsection{Component breakdown}

The reported $0.7\%$ controller overhead uses encoder latency as its denominator and covers the register query, budget readout, and top-$k$ bookkeeping. The sparse-encoder measurement includes the candidate-key projection at pruning blocks, together with the retained-token projections, attention, and MLP computation. End-to-end throughput also contains stand-in matching, pointer maintenance, sparse packing, dense reconstruction, and the task head. Table~\ref{tab:latency-breakdown} reports each component as a fraction of end-to-end latency.

The breakdown separates savings accumulated across later sparse blocks from costs paid near a pruning layer or at the dense output boundary. This distinction explains why encoder throughput reflects token reduction more directly than end-to-end throughput.

\begin{table}[t]
\centering
{\footnotesize
\setlength{\tabcolsep}{3pt}
\begin{tabular}{@{}lrrr@{}}
\toprule
Component & Cls & Seg & Det \\
\midrule
Register query/readouts & 0.0008 (0.2) & 0.006 (0.1) & 0.08 (0.1) \\
Ranking and indices & 0.0022 (0.5) & 0.019 (0.3) & 0.29 (0.3) \\
Stand-in matching & n/a & 0.09 (1.6) & 1.40 (1.6) \\
Sparse packing & 0.010 (2.2) & 0.06 (1.1) & 1.00 (1.2) \\
Sparse encoder & 0.422 (94.8) & 3.28 (58.8) & 49.52 (56.9) \\
Dense reconstruction & n/a & 0.12 (2.2) & 0.90 (1.0) \\
Task head & 0.010 (2.2) & 2.00 (35.8) & 33.77 (38.8) \\
End-to-end & 0.445 (100) & 5.58 (100) & 86.96 (100) \\
\bottomrule
\end{tabular}
}
\caption{TAP-J component latency in milliseconds per image at $\rho=0.5$, with the end-to-end percentage in parentheses. Percentages may sum to 99.9 because of rounding.}
\label{tab:latency-breakdown}
\end{table}

\FloatBarrier
\subsection{Peak memory}

Table~\ref{tab:memory} reproduces the main full-model memory values. Every measurement uses batch size 1, the task resolution in Table~\ref{tab:training-config}, and the same BF16 inference path as the latency experiment. We reset the CUDA peak-memory counter after warm-up and report the maximum allocated memory over 50 iterations. The reported value includes sparse packing, pointers, offsets, dense reconstruction, and the task head.

Peak memory does not follow encoder throughput exactly. Pruning shortens the tensors kept by later transformer blocks, but the dense pipelines must also retain indices and recovery data until their heads consume a complete grid. Segmentation benefits most because several encoder blocks operate on the shortened sequence while its recovery buffers remain small relative to the backbone activations. Detection retains larger head and feature-pyramid allocations, which limits the full-model reduction even though sparse attention removes substantial encoder work. The complete inference profile assigns controller state, recovery buffers, and head allocations to their corresponding components.

\begin{table}[H]
\centering
\begin{tabular}{lrrrr}
\toprule
Task & Base & TAP-J & TAP-F & Change J/F \\
\midrule
Cls & 0.42 & 0.37 & 0.38 & $-11.9/-9.5\%$ \\
Seg & 0.78 & 0.64 & 0.66 & $-17.9/-15.4\%$ \\
Det & 3.10 & 2.82 & 2.88 & $-9.0/-7.1\%$ \\
\bottomrule
\end{tabular}
\caption{Peak full-model inference memory in GB at $\rho=0.5$, measured with batch size 1 using the CUDA peak allocated-memory counter.}
\label{tab:memory}
\end{table}

\section{Spatial Retention Analysis}
\label{sec:spatial-retention}

The visual examples in main-paper Figure 3 suggest that the tasks preserve different regions. We test this pattern over the full validation sets at $\rho=0.5$. On ADE20K, a boundary patch contains more than one non-void ground-truth label, and boundary retention is the fraction of these patches that survive. We assign each patch its majority non-void label, then average the fraction of classes represented by at least one retained patch in each image. On COCO, box-interior retention is the fraction of patch centres inside the union of the ground-truth boxes that survive. Following the COCO area threshold, a small object covers fewer than $32^2$ pixels. Small-object coverage is the fraction of eligible objects that contain at least one retained patch centre.

\begin{table}[H]
\centering
{\footnotesize
\begin{tabular}{@{}lrrrr@{}}
\toprule
& \multicolumn{2}{c}{ADE20K} & \multicolumn{2}{c}{COCO} \\
\cmidrule(lr){2-3}\cmidrule(lr){4-5}
Method & Boundary & Class & Box interior & Small object \\
\midrule
Random & 50.0 & 92.1 & 50.0 & 50.2 \\
Attention & 44.8 & 88.7 & 62.3 & 60.8 \\
Token Cropr & 55.1 & 95.0 & 64.2 & 66.4 \\
TAP-J & 58.4 & 96.2 & 66.1 & 69.7 \\
\bottomrule
\end{tabular}
}
\caption{Spatial retention statistics in percent at $\rho=0.5$. Boundary and box-interior columns measure retained patch fractions. Class and small-object columns measure coverage.}
\label{tab:spatial-statistics}
\end{table}

\begin{figure}[H]
\centering
\includegraphics[width=0.98\columnwidth]{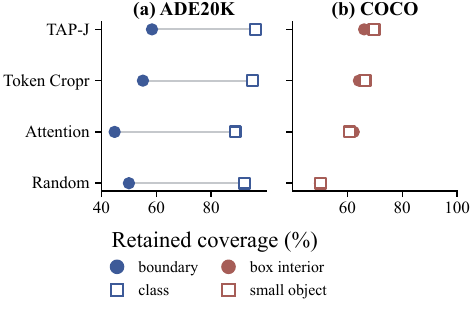}
\caption{Spatial retention statistics from Table~\ref{tab:spatial-statistics}. Filled circles measure retained boundary or box-interior patches. Open squares measure semantic-class or small-object coverage. The segmentation and detection colors match the main-paper figures.}
\label{fig:spatial-retention}
\end{figure}

Table~\ref{tab:spatial-statistics} separates local retention from coverage. Attention concentrates on box interiors in detection but loses both ADE20K boundaries and semantic classes. Token Cropr improves all four measures. TAP-J raises boundary retention by 3.3 points and class coverage by 1.2 points over Token Cropr. Its gains on COCO are 1.9 points for box interiors and 3.3 points for small objects. Figure~\ref{fig:spatial-retention} shows simultaneous gains in local evidence and coverage on both datasets.

\FloatBarrier
\section{Scope and Reproducibility Map}
\label{sec:scope}

The controlled probes compare complete task pipelines whose objectives, resolutions, heads, and eligible pruning layers differ. They measure transfer across these pipelines rather than task semantics in isolation. TAP uses a fixed user-specified keep rate within each run and does not learn the total compute budget per image. Recovery combines one stand-in with an earlier-layer offset and has no explicit transport map to the dense read point. Classification does not use this mechanism.

Detection updates the register only at global-attention blocks. Its block-3 allocation and recovery scale depend on the task but not the image, although the current patch keys still make token scores image-dependent. Later allocations and scales use image context from the preceding global block while patch states evolve through the intervening window-attention blocks. Efficient execution also requires a variable-length window-attention kernel. The reported claims apply to this global-block update schedule and variable-length execution path.

\begin{table}[t]
\centering
{\footnotesize
\begin{tabular}{@{}p{0.31\columnwidth}p{0.63\columnwidth}@{}}
\toprule
Claim & Evidence and reproduction artifact \\
\midrule
Criterion rankings change across pipelines & Main Figure 1(c)\newline Full criterion and rank tables\newline Probe configurations and script \\
Depth changes criterion quality & Main Figure 1(a,b)\newline Complete layerwise curves\newline Layer-probe script \\
Dense tasks prefer different recovery endpoints & Main Figure 1(d) and Table 4\newline Paired recovery table\newline Recovery configurations \\
The task register controls three decisions & Main Method, Table 4, and Figure 5\newline Algorithms S1--S3 and distributions\newline Model and ablation configurations \\
The global removal budget is exact & Main Method\newline Proposition and unit tests\newline Allocation tests \\
TAP balances score and throughput & Main Tables 1--3 and Figure 4\newline Keep-rate and latency tables\newline Profiler and logs \\
TAP lowers peak inference memory & Main Table 5\newline Memory protocol and breakdown\newline Memory profiler \\
\bottomrule
\end{tabular}
}
\caption{Mapping from central claims to the main paper, supplemental evidence, and reproduction artifacts. Each evidence cell lists these three sources in order.}
\label{tab:claim-map}
\end{table}

The anonymous artifact records the environment, data preparation, configurations, seeds, probe and profiling scripts, checkpoint checksums, and commands for each reported result. The configuration record includes optimizer coefficients, numerical precision, layerwise learning-rate decay, stochastic-depth settings, gradient clipping, loss weights, augmentation ranges, and the checkpoint-selection rule. The artifact omits author names, institutional paths, and identifying metadata.

\bibliography{aaai2027}